\documentclass[pdflatex,sn-mathphys-num]{sn-jnl}
\usepackage{graphicx}
\usepackage{multirow}
\usepackage{amsmath,amssymb,amsfonts}
\usepackage{amsthm}
\usepackage{mathrsfs}
\usepackage[title]{appendix}
\usepackage{xcolor}
\usepackage{textcomp}
\usepackage{manyfoot}
\usepackage{booktabs}
\usepackage{algorithm}
\usepackage{algorithmicx}
\usepackage{algpseudocode}
\usepackage{listings}
\usepackage{makecell}
\usepackage{svg}

\begin{document}

\title[Adaptive MFML]{Improvise, Adapt, Overcome: An On-The-Fly Multifidelity Algorithm for Efficient Machine Learning}

\author*[1]{\fnm{Vivin} \sur{Vinod}}\email{vinod@uni-wuppertal.de}
\author[1]{\fnm{Peter} \sur{Zaspel}\email{zaspel@uni-wuppertal.de}}
\affil[1]{\orgdiv{School of Mathematics and Natural Sciences}, \orgname{University of Wuppertal}, \orgaddress{\country{Germany}}}

\abstract{
Machine learning has accelerated quantum chemistry but is hindered by the prohibitive cost of generating high fidelity training data. Multifidelity machine learning (MFML) mitigates this overhead by systematically combining abundant low fidelity data with sparse high fidelity data. In spite of its success, standard MFML schemes rely on pre-defined scaling factors to determine sparse data ratio across fidelities, often generating redundant multifidelity data resulting in a loss of efficiency. Here, we introduce an adaptive on-the-fly multifidelity framework for machine learning that autonomously determines training dataset composition. By dynamically querying training samples at each fidelity, the algorithm saturates model accuracy at lower fidelities before moving up to more expensive reference calculations. We benchmark the novel adaptive-MFML across diverse chemical properties including theoretical best estimate \textit{ab initio} potential energy surfaces, golden standard coupled cluster energies, and the more chemically challenging excitation energies. In our numerical experiments we show that our adaptive algorithm reduces data generation costs by up to a factor of 20 compared to single fidelity methods and improves upon standard MFML by up to a factor of 3. The mitigation of data redundancy establishes a high-accuracy low-cost pathway for sustainable cost-aware machine learning in quantum chemistry. 
}

\maketitle

\section{Introduction}
\label{sec:introduction}
In computational simulations, increasing the cost-budget typically yields higher accuracy, introducing the concept of \textit{fidelities} relative to a ground truth \cite{Kennedy_2000_MF_fastapprox_copmutercodes, Gratiet_Garnier_2014}. The cost-accuracy tradeoff is central in multifidelity methods where the aim is to achieve high-accuracy low-cost models by strategically combining outputs across fidelities. These methods often utilize data-driven surrogates, such as machine learning (ML) models, to \textit{comprehensively correct} a low fidelity (computationally cheap) model to match the accuracy of the higher fidelity model at a much lower cost than running the higher fidelity model. Multifidelity methods have seen a wide range of applications ranging from geological sciences to fluid simulations \cite{Fernandez-Godino_2023}.

The use of ML has significantly accelerated research in quantum chemistry (QC), replacing costly calculations with accurate predictions \cite{dral2020quantum, westermayr_2021_perspective}. However, these surrogate models require substantial training data and are usually restricted to the fidelity of the training data. This caveat has been of some discussion only in the recent past, with increasing notice being paid to this computational bottleneck. High accuracy reference data, such as the gold standard of Coupled Cluster with Singles, Doubles, and Perturbative Triples (CCSD(T)), scales prohibitively with system size as $\mathcal{O}(N^7)$ \cite{Crawford_CCSD_theory_2000} severely limiting the ML-QC workflow for high fidelity QC properties. 

Multifidelity machine learning (MFML) has emerged as a prominent solution to the training data cost bottleneck. Following the \textit{ansatz} of standard multifidelity methods \cite{Kennedy_2000_MF_fastapprox_copmutercodes,Gratiet_Garnier_2014}, the comprehensive correction of low fidelity ML models towards high fidelity QC predictions has been used successfully to predict properties ranging from ground state energies to excitation energies and bandgaps of solids showing significant reduction in the time-cost of the ML-QC pipeline \cite{Ramakrishnan2015, Pilania2017, zasp19a, vinod23_MFML, Ruth_delML_CCSD_DFT, Vinod2024_multifidelity_monomers_CCSDT}. With increasing interest in sustainable chemical discovery and reducing computation footprint of ML in QC \cite{sandonas2026perspectivesustainableexplorationchemical}, MFML methods have increasingly become popular not only for their prediction accuracy but also for their ability to reduce the computation cost incurred \cite{dral2020hierarchical, vinod23_MFML, vinod2024_benchmarking_dataefficiency_deltaml}. 
While the MFML methods are well established for small, isolated molecules \cite{dral2020hierarchical, vinod2024_gamma_curve_error_contours}, they also show great promise for scalability in extended molecular systems. Recent applications highlight this capability, ranging from simulating exciton energy transfer in a collection of 85-atom porphyrins on a clay surface for synthetic light-harvesting \cite{lyu2024_porphyrin} to over 3.5 million point predictions of high fidelity excitation energies in a light harvesting complex of photosynthetic bacteria \cite{vinod_2026_deltaBCL}. Further evidence in recent works indicates that MFML methods exhibit data-efficiency which supersede $\Delta$-ML schemes in large number of prediction regimes \cite{vinod2024_benchmarking_dataefficiency_deltaml}.  

In spite of their success, standard MFML methods address data sparsity with increasing fidelity by relying on a fixed scaling factor to determine the ratio of training samples between two fidelities. Conventionally, the scaling factor is set to 2 \cite{zasp19a, vinod23_MFML} derived from legacy of sparse grid combination techniques \cite{Bungartz_Griebel_2004}. Other fixed values of the scaling factor have been explored indicating that larger values improve the cost-benefit effectiveness of the MFML models \cite{vinod2024_gamma_curve_error_contours, lyu2024_porphyrin}. An alternate approach for the use of multifidelity methods in ML-QC introduced an \textit{ad-hoc} heuristic to dictate training data set sizes across fidelities omitting the fixed scaling factor \cite{dral2020hierarchical}. The scheme, hierarchical ML,  uses several $\Delta$-ML models \cite{Ramakrishnan2015} and assumes a constant ratio of predictions between two $\Delta$-ML models for subsequent fidelities remains approximately constant. This scheme was shown to be effective in reducing training data costs for a specific isolated small molecule by a factor of 100. In such a setup, a noted constraint is that the constant ratio assumption breaks down for small training set sizes and requires a \textit{post-hoc} optimization. Further, since the error is estimated, an \textit{a-posteriori} calibration of the final trained model could be required. Ultimately, both fixed scaling factor MFML and \textit{ad-hoc} estimated training dataset sizes fail to dynamically capture the true cost-benefit contribution of each fidelity during training the MFML model. Inevitably, this leads to an increased risk of redundancy of costly training data generation requiring manual \textit{post-hoc} intervention.

With cost-aware ML gaining traction in QC, there is a pressing need to develop robust adaptive schemes for the dynamic selection of optimal number of training samples per fidelity. Previously, the use of a greedy-framework to address multifidelity training data set sizes using cost-sensitive information acquisition reducing posterior uncertainty of the target function has been proposed as an approach \cite{pmlr_v89_song19b, Phan_Byun_2025}. In these approaches, Gaussian Processes (GPs) are utilized to model the target properties, and probabilistic acquisition functions evaluate the design space to select the next optimal individual data point.
Elsewhere, the use of deep neural networks in the same ansatz resulted in efficient multifidelity models for the prediction of vector quantities for fluid dynamics \cite{pmlr_v151_li22b}. In this case, variational inference was used to inform training data acquisition at each fidelity.
Closer to the application targeted in this current work, Heinen \textit{et al.}~introduced the M3L scheme \cite{Heinen_Khan_Falk_Lilienfeld_2024} which utilizes a Bayesian Optimization scheme that learns the relation between number of training samples and model prediction error to find optimal ratios of training samples between fidelities. 

The above mentioned approaches have indeed begun to address the \textit{a priori} composition of training dataset. However, such schemes bypass fixed heuristics by introducing complex probabilistic acquisition functions across chemical design spaces producing further computational strain. Alternatively, the use of cost-optimized frameworks such as M3L ultimately relies on fixed dataset ratios which results in a gap for a computationally light-weight algorithm that carries out on-the-fly adaptive assessment of model error. 
Such an algorithm would not only minimize expensive training data generation but also relies on an on-the-fly computation of reference QC properties. Unlike conventional MFML approaches that require \textit{a-priori} datasets across all fidelities, a truly adaptive sampling scheme queries QC reference calculations strictly on a need-to-know basis eliminating redundancy in the ML-QC pipeline by generating only the data necessary to improve the model. 

This work introduces a novel adaptive-MFML scheme that performs on-the-fly sampling of training data at each fidelity to minimize the empirical error of the resulting MFML model. Recursively adding training samples at each fidelity, the algorithm ensures the error reduction meets a user-defined threshold. We demonstrate in the following section that this new adaptive algorithm reduces the computation cost of ML-QC methods by an order of magnitude compared to single fidelity methods, and improves on basic MFML baselines by a factor of 3 in some cases. The algorithm is shown to be robust and reduce redundancy. The applicability of the proposed adaptive-MFML framework across diverse QC properties is demonstrated by predicting properties ranging from \textit{ab initio} potential energy surfaces (PES) to more complex and challenging excitation energies.

\section{Methods}\label{sec:methods}
This section describes the online methods used in this work. Details regarding Kernel Ridge Regression (KRR) architecture, dataset composition, and molecular descriptor generation are provided Supplementary Information sections S1--S3.

\subsection{Multifidelity Machine Learning}\label{sec:MFML}
Let an ordered set of fidelities $f=\{1,2,\ldots,F\}$ be given such that the accuracy of the evaluation of the computation (the oracle), $g^f(\boldsymbol{x})=y^f$ is greater than the accuracy of $g^{f-1}(\boldsymbol{x})=y^{f-1}$. Thus, $\mathcal{T}^f:=\left\{(\boldsymbol{x}_i,y_i^f)\right\}_{i=1}^{N_f}$ represents the training set at fidelity $f$. For all experiments carried out in this work, we use mean-centered outputs $y$. 
A recursive comprehensive correction of fidelities formulated for unseen input features $\boldsymbol{x}_q$ as
\begin{equation}
    P^f(\boldsymbol{x}_q):= \rho_{f-1}(\boldsymbol{x}_q)P^{f-1}(\boldsymbol{x}_q) + \delta_f(\boldsymbol{x}_q)~,
    \label{eq_comprehensive_correction}
\end{equation}
where $P^f$ is the output from a surrogate model for fidelity $f$, $\rho$ is the multiplicative corrective term, and $\delta$ is the additive correction term \cite{Gratiet_Garnier_2014}. 
Multifidelity machine learning uses this concept and generalizes it to train sub-models at different fidelities \cite{zasp19a, vinod_2024_oMFML} . Each sub-model is identified by the fidelity $f$ and the number of training samples $N$. That is, a composite index $\boldsymbol{s}:=(f,N)$ is used. The number of training samples halves at each successive fidelity, that is, $N_{f-1}=2\cdot N_{f}$. 
Since data sparsity increases with increase in fidelity of data, MFML requires a nested training data structure \cite{vinod2024_nonnestedMFML}. Define $\mathcal{X}^f:=\left\{\boldsymbol{x}_i\lvert (\boldsymbol{x}_i,y_i^f)\in\mathcal{T}^f\right\}$. The nestedness of multifidelity data is then expressed as:
\[
\mathcal{X}^F\subseteq\mathcal{X}^{F-1}\subseteq\ldots\subseteq\mathcal{X}^1~.
\]

For a query descriptor $\boldsymbol{x}_q$ the final high fidelity prediction is the linear combination of sub-model predictions:
\begin{equation}
    P_{\rm MFML}(\boldsymbol{x}_q):= \sum_{\boldsymbol{s}\in\mathcal{S}}\beta_{\boldsymbol{s}}P_{\rm ML}^{\boldsymbol{s}}(\boldsymbol{x}_q)~,
    \label{eq_MFML}
\end{equation}
where $\mathcal{S}$ is the set of all sub-models selected to be part of the MFML structure \cite{vinod_2024_oMFML}. For a given baseline fidelity ($f_b$), that is the lowest fidelity to be used in the MFML model, target fidelity ($F$), and number of training samples at the target fidelity ($N_F$), one can write:
\begin{align*}    
    \mathcal{S} :=  \Big{\{}(f,N_F) &\Big{\lvert}
    f \in\left\{f_b,\ldots,F\right\},\\
     &N_F\in\{N_F,\ldots,2^{F-f_b}\cdot N_F \},\\
    &F+N_F-1\leq f+N_F \leq F+N_F \Big{\}}
    \label{eq_subspace_MFML}~,
\end{align*}
The coefficients $\beta_{\boldsymbol{s}}$ are assigned unitary signed values ($\pm1$). In this work, we utilize a KRR architecture for the sub-models $P^{\boldsymbol{s}}_{\rm ML}$. However, it must be noted that the MFML scheme is architecture-agnostic and can be implemented for alternate architectures such as neural networks \cite{Pilania2017}.

As an example, consider a three fidelity hierarchy, that is, $f=\{1,2,3\}$ with $N_F=4$. Then, the prediction from the MFML model using the KRR architecture for the sub-models is given as 
\begin{align*}
P_{\rm MFML}(\boldsymbol{x}_q)&:= P_{\rm KRR}^{(f=3,4)}(\boldsymbol{x}_q)-P_{\rm KRR}^{(f=2,4)}(\boldsymbol{x}_q)\\
&+P_{\rm KRR}^{(f=2,8)}(\boldsymbol{x}_q)-P_{\rm KRR}^{(f=1,8)}\\
&+P_{\rm KRR}^{(f=1,16)}(\boldsymbol{x}_q)~.
\end{align*}
This expansion reveals that the prediction of an MFML model can be seen as a telescopic sum across fidelities:
\begin{equation}
    P_{\rm MFML}(\boldsymbol{x}_q):= \sum_{f=1}^F\left[P_{\rm KRR}^{(f,N_f)}(\boldsymbol{x}_q) - P_{\rm KRR}^{(f-1,N_f)}(\boldsymbol{x}_q)\right]~,
    \label{eq_telsum_MFML}
\end{equation}
where we define $P_{\rm KRR}^{(0,N)}(\boldsymbol{x}_q)\equiv0$ explicitly.

\subsection{Adaptive Multifidelity Sampling}\label{sec:adaptive}

\begin{algorithm}[!htbp]
\caption{Adaptive Multifidelity Sampling algorithm for an ordered hierarchy of fidelities $f\in\{1,2,\ldots,F\}$. Let $E_k$ be the validation MAE after the $k$-th batch addition and let $\bar{E}_k:=\frac{1}{w}\sum_{j=k-w+1}^{k}E_j$ be its moving average over a window of $w$ iterations. The \texttt{Local\_Improvement} is the absolute change of this moving average, $\Delta_k:=|\bar{E}_k-\bar{E}_{k-1}|$. The \texttt{Global\_Improvement} is the reduction of the validation MAE over one complete pass through all fidelities, $\Delta_{\rm global}:=E^{(p-1)}-E^{(p)}$, where $E^{(p)}$ is the MAE at the end of global pass $p$.}
\label{alg:adaptive_mfml}
\begin{algorithmic}[1]
\Require initial training set sizes $\boldsymbol{N_{\rm train}}:=[N_1,N_2,\ldots,N_F]$, 
batch sizes $\boldsymbol{m_{\rm batch}}:=[m_1,m_2,\ldots,m_F]$, 
local tolerances $\boldsymbol{\tau}:=[\tau_1,\tau_2,\ldots,\tau_F]$, 
global tolerance $\tau_{\rm global}$,
validation dataset $\mathcal{V}^F$, pool of unlabeled geometries $\mathcal{X}_{\rm pool}$
\Ensure QC calculator $g^f_{\rm QC}~,\forall f\in\{1,\ldots,F\}$    
\State $\mathcal{T}^f\gets \left\{(x_i,g^f_{\rm QC}(x_i))\right\}_{i=1}^{N_f}~, \forall f\in\{1,\ldots,F\}$ 
\State $\mathcal{T}_{\rm MFML} \gets \left[\mathcal{T}^1,\ldots,\mathcal{T}^F\right]$ \Comment{Initial multifidelity dataset}
\State $E \gets \text{MAE}(\mathcal{T}_{\rm MFML},\mathcal{V}^F)$
\State $E_{\rm prev} \gets E$
\State $\Delta_{\rm global}\gets E$ \Comment{Force at least one global pass}
\While{$\Delta_{\rm global}>\tau_{\rm global}$} \Comment{Repeat the sampling scheme}
\For{$f = 1$ \textbf{to} $F$} \Comment{Step up through each fidelity}
    \State $E \gets \text{MAE}(\mathcal{T}_{\rm MFML},\mathcal{V}^F)$
    \State $\Delta \gets E$ \Comment{Re-initialise local statistic (enter loop if $E>\boldsymbol{\tau}[f]$)}
    \While{$\Delta>\boldsymbol{\tau}[f]$} \Comment{Sample at fidelity $f$}
        \If{$f>1$ and $2\times(\boldsymbol{N_{\rm train}}[f] + \boldsymbol{m_{\rm batch}}[f]) \ge \boldsymbol{N_{\rm train}}[f-1]$}
            \State \textbf{break} \Comment{Maintain hierarchical size constraint}
        \EndIf
        \If{$f>1$}
            \State $\mathcal{X}^f\gets \{x:(x,y)\in\mathcal{T}^{f-1}\}\setminus\mathcal{E}^f$     \Comment{Ensure nested MF data, each geometry queried once} \Comment{Ensure nested MF data}
        \EndIf
        \State $\mathcal{T}^f_{\rm batch}\gets \left\{(\boldsymbol{x}_i,g^f_{\rm QC}(\boldsymbol{x}_i))\right\}_{i=1}^{\boldsymbol{m_{\rm batch}}[f]} \text{, s.t.}~x_i\in\mathcal{X}^f$ \Comment{\textbf{On-the-fly computations}}
        \State remove $\{\boldsymbol{x}_i\}$ from $\mathcal{X}_{\rm pool}$ \Comment{Each geometry is queried once}
        \State $\mathcal{T}^f\gets\mathcal{T}^f\cup\mathcal{T}^f_{\rm batch}$ \Comment{Add samples at current fidelity}
        \State $\mathcal{E}^f\gets\mathcal{E}^f\cup\{\boldsymbol{x}_i\}$ \Comment{Update evaluated set}
        \State $\mathcal{T}_{\rm MFML} \gets \left[\mathcal{T}^1,\ldots \mathcal{T}^f,\ldots,\mathcal{T}^F\right]$ \Comment{Update multifidelity training data}
        \State $N_{\rm train}[f] \gets \boldsymbol{N_{\rm train}}[f] + \boldsymbol{m_{\rm batch}}[f]$ \Comment{Update training set size}
        \State $E \gets \text{MAE}(\mathcal{T}_{\rm MFML},\mathcal{V}^F)$
        \State $\Delta \gets |\bar{E}_k-\bar{E}_{k-1}|$ \Comment{Local improvement}
    \EndWhile
\EndFor
\State $\Delta_{\rm global}\gets E_{\rm prev}-E$ \Comment{Global improvement}
\State $E_{\rm prev}\gets E$
\EndWhile
\State \Return $\boldsymbol{N_{\rm train}}, \mathcal{T}^f, \forall f\in\{1,2,\ldots,F\}$
\end{algorithmic}
\end{algorithm}
The MFML scheme described above has been successfully used to reduce training data costs for various QC properties such as excitation energies \cite{vinod23_MFML, vinod_2024_oMFML} and atomization energies \cite{zasp19a}. However, in the standard MFML model, the number of training samples at each fidelity is decided by heuristics motivated by sparse grid combination techniques \cite{Bungartz_Griebel_2004} resulting in the rigid relation $N_{f-1}=2\cdot N_{f}$. 
The main contribution of this work is an algorithm that adaptively samples the number of training samples needed for each fidelity on-the-fly. This procedure minimizes redundancy in training data generation, lowering the total computational footprint of MFML for QC. 

Let a validation set at the highest fidelity be given as $\mathcal{V}^F:=\left\{(\boldsymbol{x}_i,y_i^F)\right\}_{i=1}^{N_{\rm val}}$. In standard ML practices this dataset is used to optimize hyper-parameters of the ML model (e.g.: kernel, kernel width etc). 
In our adaptive-MFML framework, the training data set size at each fidelity, $N^f$, is treated as a dynamic variable optimized using the validation dataset. As illustrated by the workflow in Fig.~\ref{fig_concept}, the optimization is carried out within a nested loop structure. For each local loop (or epoch), the algorithm sequentially injects a batch of training data starting at the lowest fidelity ($f=1$). As presented in Algorithm \ref{alg:adaptive_mfml}, in each epoch, an MFML model is trained and evaluated for $\mathcal{V}^F$ computing the empirical error. Let $E_k$ denote the validation MAE after the $k$-th batch addition and $\bar{E}_k$ its moving average over a window of $w$ iterations. The local improvement tracked by the algorithm is $\Delta_k=|\bar{E}_k-\bar{E}_{k-1}|$, that is, the change of the windowed average between consecutive iterations. On entering each fidelity loop, $\Delta$ is re-initialized to the current MAE, so the loop is entered whenever the current validation error still exceeds the tolerance. This prevents the termination of one fidelity loop from suppressing the subsequent ones when their tolerances are equal. If $\Delta$ falls below a \textit{local tolerance} threshold, the algorithm moves to the next fidelity and carries out the same procedure. This step-wise progression is performed until the highest fidelity is reached. At this stage, the algorithm checks for improvement in the empirical error for one global loop (that is cycling through all fidelities once). If this \textit{global improvement} is below a \textit{global tolerance}, the algorithm terminates returning returning the adaptively sampled training set sizes and a trained MFML model. Otherwise, the algorithm continues to dynamically add training samples starting from the lowest fidelity passing through epochs and global loops.
Notice that for each fidelity, during an epoch, the algorithm calls a QC calculator at the on-the-fly computation step of Algorithm~\ref{alg:adaptive_mfml}. That is, the algorithm only requires the calculator to compute reference computations on a need-to-know basis. Further, the hierarchical size-constraint step of Algorithm~\ref{alg:adaptive_mfml} terminates the local loop if the ratio of training dataset sizes for consecutive fidelities reaches the
standard MFML fixed scaling of 2. At the same time, since the training data are added at the lowest fidelity first, the training data at the higher fidelities are only added from the subset of geometries that are adaptively selected at the lowest fidelity.

To make both the nestedness and the uniqueness of the reference computations explicit, the implementation maintains, for each fidelity, the set of already-evaluated geometries
\[
    \mathcal{E}^f=\{x:(x,y)\in\mathcal{T}^f\}.
\]
The candidate pool at fidelity $f$ is then
\[
    \mathcal{X}^f=\{x:(x,y)\in\mathcal{T}^{f-1}\}\setminus\mathcal{E}^f
    \qquad\text{for } f>1,
\]
and, for $f=1$, the prescribed base design pool $\mathcal{X}_{\rm pool}$ minus $\mathcal{E}^1$. The batch is sampled uniformly \textit{without replacement} from $\mathcal{X}^f$,
\[
    \mathcal{T}^f_{\rm batch}=\{(x_i,g^f_{\rm QC}(x_i))\}_{i=1}^{m_f},
    \qquad x_i\in\mathcal{X}^f,\quad x_i\neq x_j\ \text{for } i\neq j,
\]
after which $\mathcal{E}^f$ is updated. If $|\mathcal{X}^f|<m_f$, the implementation reduces the batch size to $|\mathcal{X}^f|$ and flags candidate exhaustion. It does not resample already evaluated geometries at the same fidelity.
By construction, a geometry is therefore never selected twice at the same fidelity. For $f>1$ the candidate pool consists of the geometries labeled at fidelity $f-1$ with those already labeled at $f$ removed, while at $f=1$ every selected geometry is removed from $\mathcal{X}_{\rm pool}$. The same geometry may of course appear at lower and higher fidelities, which is the intended nested multifidelity construction. This guarantees $\mathcal{X}^F\subseteq\mathcal{X}^{F-1}\subseteq\ldots\subseteq\mathcal{X}^1$ while keeping every reference computation unique.

\subsection{Error Metrics}\label{sec_errormetrics}
The prediction error of ML models studied in this work are computed as the mean absolute error (MAE) over a holdout test dataset $\mathcal{U}^F:=\{(\boldsymbol{x}_i,y^F_i)\}_{i=1}^{N_{\rm test}}$ which is given as 
\begin{equation}
    \mathrm{MAE}:= \frac{1}{N_{\rm test}}\sum_{i=1}^{N_{\rm test}}\left\lvert y_i^F-P_{\rm ML}(\boldsymbol{x}_i)\right\rvert~.
    \label{eq_mae}
\end{equation}
For the ANI-1ccx dataset, instead of considering MAE, we use the relative MAE (RMAE), computed as follows:
\begin{equation}
    \mathrm{RMAE}:= \frac{1}{N_{\rm test}}\sum_{i=1}^{N_{\rm test}}\left\lvert \frac{y_i^F-P_{\rm ML}(\boldsymbol{x}_i)}{y^F_i}\right\rvert~.
    \label{eq_rmae}
\end{equation}
At the same time, in some visualization, we present the relative error per molecule. This is computed as 
\[
\text{Error}_{rel,i}:=\left(\frac{y_{i}^F - P_{\rm ML}(\boldsymbol{x}_i)}{y_{i}^F}\right)
\]
The test set is not used at any stage of training or hyper-parameter optimization to truly reflect an out-of-sample error metric. It is to be noted that the test set is not intended to be a leave-one-out strategy, that is, evaluating for transferability. Rather, keeping with common practice in assessing ML methods, it is a set of data points that are never included in the training or validation phase of the ML models. 

In this study, the MAE of the trained models are studied as a function of the training data cost. For a single fidelity model the cost of a model is $T_{\rm train}^{\rm SF}:=N_F\cdot c_F$ where $c_F$ is the cost of a single data point at fidelity $F$. For the multifidelity models, the training data cost is the cumulative cost of training data used at each fidelity, that is, $T_{\rm train}^{\rm MFML}:= \sum_{f=1}^FN_f\cdot c_f$. For the basic MFML model, due to the data scaling, this reduces to $T_{\rm train}^{\rm MFML}=\sum_{f=1}^F 2^{f-1}N_1\cdot c_f$. As an example, consider the case for $N_{\rm train}=[32,16,8,4]$ for the ANI-1ccx dataset. The time-costs are reported in Table ST3 in the supplementary information. The cost of the MFML with this specific training data structure will be $32\cdot9.65+16\cdot16.82+8\cdot40.35+4\cdot157.27=1529.8$ seconds. 

For the adaptive-MFML scheme, one additionally requires high-fidelity reference labels for the validation set to track the error improvements. Thus the total acquisition cost of a run is
$T := T_{\rm train} + N_{\rm val}\,c_F$. Across every dataset and every experiment reported here the validation set is fixed at the smallest practical size, $N_{\rm val}=2$, that is, two highest-fidelity reference computations per run (for ANI-1ccx this adds
$2\times157.27~{\rm s}\approx0.09$~h). Single-fidelity KRR and basic MFML require no validation set and are therefore reported at training cost only. A study of the validation-set size, $N_{\rm val}\in\{2,4,8,16,32\}$, together with the corresponding validation costs, is given in section S10 and table ST6 of the SI. Therein we observe that the final accuracy is essentially independent of $N_{\rm val}$ over this range.

Since one is generally interested in the average behavior of the ML models, 10 independent random shuffles the training dataset before building the ML models are considered. For the singe fidelity models, this is the basic cross-validations schemes on a fixed test set. However, for the MFML models since the nested training data structure is to be retained, the cross-validation scheme introduced in \cite{vinod_2024_oMFML} is utilized to run different training dataset shuffling for the MFML models. The learning curve is then plotted as the average of the MAE resulting from these runs. For single fidelity KRR and basic MFML, the average of 10-run random shuffling of the dataset is used. For the adaptively sampled MFML models, since the random shuffling of the dataset results in a different optimization route for the training samples,a modified strategy outlined below is carried out:
\begin{enumerate}
    \item For 10 random seeds $k$ generate adaptive-MFML model $M_k$. 
    \item Collect the pair $\left(\left[N_1,N_2,\ldots,N^F\right]^k_j,\text{MAE}^k_j\right)\rightarrow\left(T_{\rm train,j}^k,\text{MAE}_j^k\right)$ for each iteration $j$ of the adaptive sampling scheme. 
    \item Build an interpolation model for $T_{\rm train}$ and MAE using the different random seed $k$. 
    \item Interpolate the average trend of the learning curves.
\end{enumerate}

\section{Results}\label{sec:results}
\begin{figure}[htb!]
    \centering
    \includegraphics[width=\linewidth]{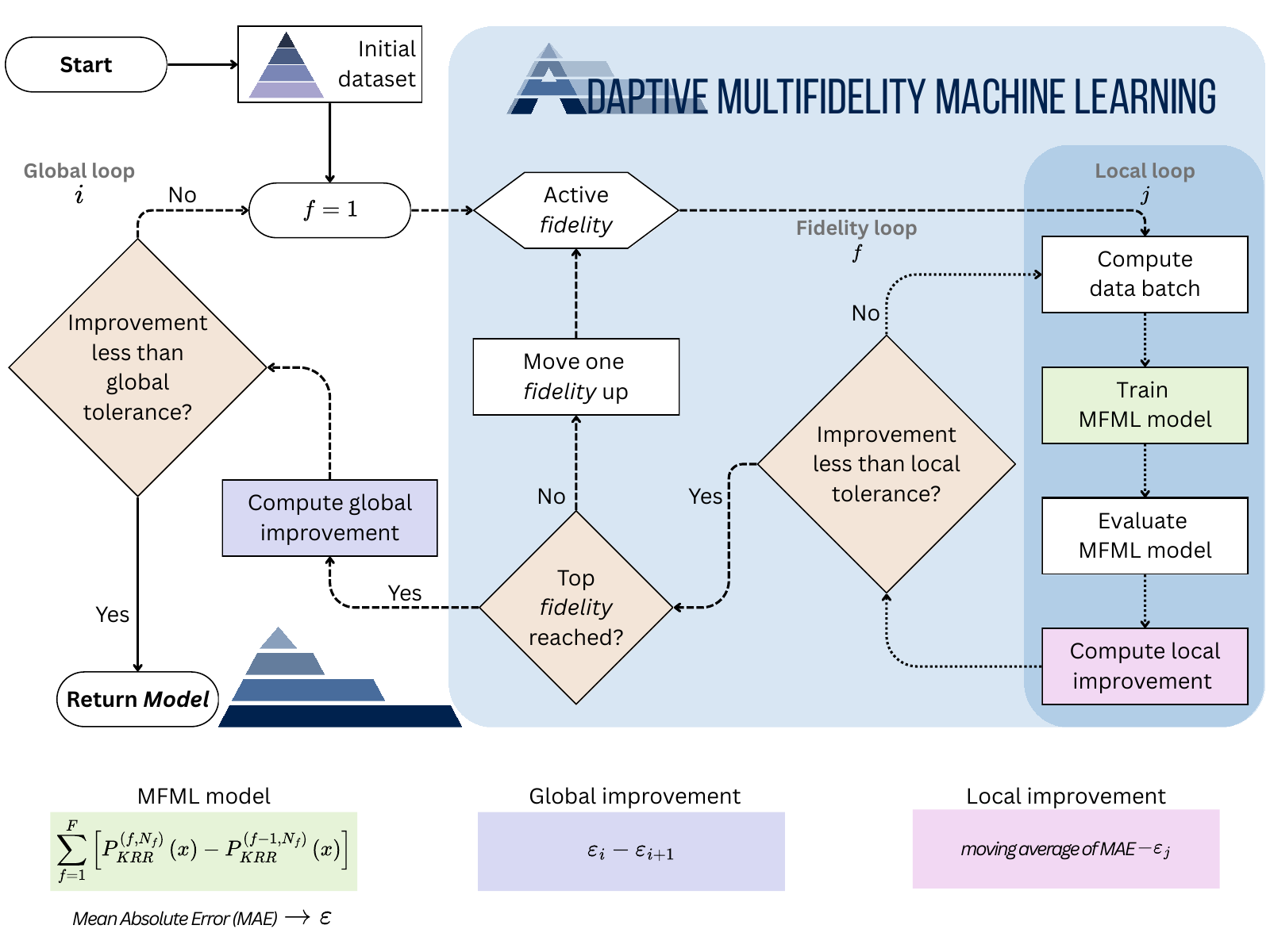}
    \caption{The conceptual workflow of adaptively selecting training samples at each fidelity. The local loop performs on-the-fly computations of QC properties for a given fidelity. Tolerance checks at each fidelity and at the global level, that is once all fidelities have been accessed once, ensure prevention of redundant training data generation.}
    \label{fig_concept}
\end{figure}

The conceptual workflow of the adaptive sampling framework for MFML models is depicted in Fig.~\ref{fig_concept}. The multifidelity training dataset is initialized using a few randomly sampled geometries with corresponding QC properties computed at varying fidelities. Across all experiments presented herein, we define four discrete fidelities, $f\in\{1,2,3,4\}$, with the initial dataset size $N_{\rm train}=[32,16,8,4]$, that is, 32 training samples at fidelity 1, 16 at fidelity 2 and so on. This reflects a small basic MFML architecture. The algorithm itself operates in a nested loop structure. In the local loop (hereon also referred to as an epoch), the algorithm dynamically adds training data starting with the lowest fidelity. During each epoch, an MFML model of the corresponding training data set composition is trained and evaluated against a small validation set of high-fidelity reference computations to keep track of the prediction error computed as the mean absolute error (MAE; see section \ref{sec_errormetrics}). To avoid pre-asymptotic artifacts that arise due to very small dataset sizes, the improvement of the MFML model is assessed over a moving average of the MAE. If the improvement falls below a predefined \textit{local tolerance}, the algorithm advances to the next fidelity to carry out on-the-fly querying of training data. The algorithm allows one to set individual local tolerance values per fidelity which can arise from domain expertise on what improvement is expected for a fidelity. For instance, it is reasonable to claim that the addition of a geometry with high fidelity reference computed should give better prediction accuracy compared to the same geometry with a low fidelity reference.

This step-wise progression of fidelities is  carried out until the highest fidelity is reached, completing one \textit{global loop}. On completing a global loop, the algorithm checks the overall error reduction across the full pass from lowest to highest fidelity. If the error has improved compared to the previous pass, the sequence restarts at the lowest fidelity. If the improvement is below a set threshold of \textit{global tolerance}, the algorithm terminates returning the adaptively sampled multifidelity training dataset and the final trained model. This form of aggressive pruning of redundant reference computation queries and growing a dataset strictly on a need-to-know basis allows the adaptive framework to avoid \textit{a-priori} generation of expensive data directly addressing the cost-budget bottleneck in the ML-QC pipeline. 

In order to establish the efficacy of the novel adaptive sampling method, we report MAE using holdout test sets that are unseen in both training and algorithmic looping. This ensures that errors reflect true generalization. A comparison of the validation set error (used to carry out sampling in the adaptive algorithm) and the holdout test set error are presented in the supplementary information Fig.~SF3 to reason that the validation set error does match the test set error indicating that the algorithm is indeed robust for generalization. The core ML architecture used in this work is Kernel Ridge Regression (KRR) motivated by its proven robustness in multifidelity QC applications \cite{zasp19a, dral2020hierarchical, vinod2024_gamma_curve_error_contours}. 

For each experiment, the total geometries are partitioned into a train:validation:test split of 85:5:10. We simulate the on-the-fly nature of the adaptive-MFML algorithm by querying the pre-computed QC reference values from the datasets only when requested in the epoch. The adaptive-MFML method is benchmarked against two baselines: a single fidelity KRR model trained only on highest fidelity data, and a standard MFML model where data sparsity is dictated by the conventional fixed scaling factor of 2 (that is, $N_{\text{train}}^{f-1}=2\cdot N_{\text{train}}^f$). 
We repeat each training with 10 independent random seeds and assess the holdout test-set prediction error of all three methods on the same fixed test set.
For MFML methods, cross validation is carried out with adherence to a nested data structure (see section \ref{sec:MFML}) requirement outlined in Ref.~\cite{vinod_2024_oMFML}. We analyze the MAE of each model as a function of the cumulative time-cost incurred in training data generation ( see section \ref{sec_errormetrics}). For single fidelity models, the cost is the product of the highest fidelity per sample cost and the training set size. For the MFML models, the cost is the sum of costs across all queried fidelities. An assessment of these time-cost learning curves provides a direct measure of data efficiency making it a critical metric in understanding the computational footprint of cost-aware ML models in QC.

\subsection{Theoretical Best Estimate of Potential Energy Surfaces}
\begin{figure}[htb!]
    \centering
    \includegraphics[width=\linewidth]{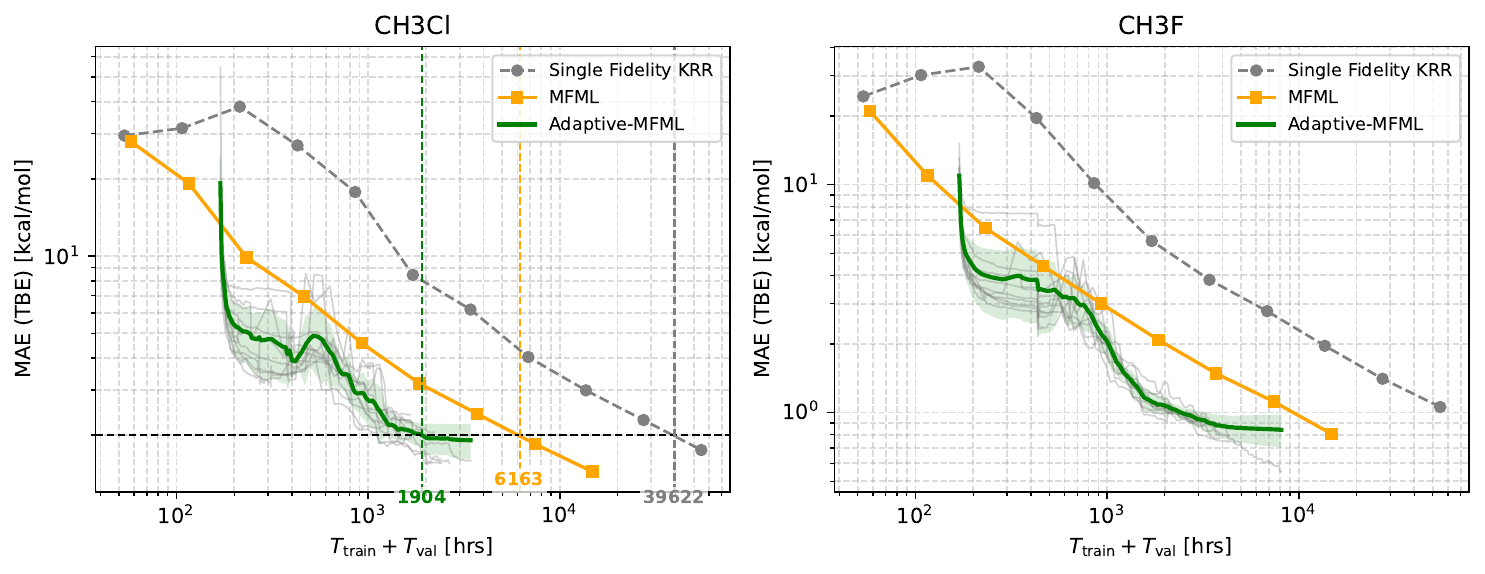}
    \caption{Time-cost of training data versus MAE for prediction of \textit{ab initio} potential energy surfaces for the TBE for $\rm CH_3Cl$ and $\rm CH_3F$ of the VIB5 dataset\cite{zhang_vib5_2022}. The adaptive-MFML model achieves the target MAE at a significantly lower computational cost than both standard MFML and single fidelity KRR.}
    \label{fig_VIB5_LC}
\end{figure}

The VIB5 dataset provides high accuracy \textit{ab initio} potential energy surfaces (PES) for $\rm CH_3Cl$ and $\rm CH_3F$ molecules \cite{zhang_vib5_2022}. For each molecule, we consider four fidelity levels, listed in increasing order of computation cost: MP2/aug-cc-pVQZ, CCSD(T)-F12b/cc-pVTZ-F12, CCSD(T)-F12b/cc-pVQZ-F12, and a Theoretical best estimate (TBE) (see Supplementary Information S3 for comprehensive details).
Fig.~\ref{fig_VIB5_LC} presents the MAE as a function of training data generation time-cost for three distinct methods: single fidelity KRR, standard MFML, and adaptive-MFML. 
For the adaptive-MFML, we report the averaged learning curve as well as the independent-run trajectories.

Consider the curves for $\rm CH_3Cl$ on the left side of Fig.~\ref{fig_VIB5_LC}. 
The horizontal and vertical dashed lines indicate the total time-cost required to reach a target MAE of $\sim$2~kcal/mol, where the cost of a run is $T=T_{\rm train}+N_{\rm val}c_F$ as defined in Section~2.3. To achieve this accuracy, single-fidelity KRR demands a total budget of $T\approx39{,}600$~h, whereas the standard MFML model requires $T\approx6{,}200$~h. 
The adaptive-MFML model reaches this target error with just $T\approx1{,}900$~h, of which 53.4~h are the $N_{\rm val}=2$ validation labels. This translates to a 20-fold improvement over single-fidelity KRR and a 3.2-fold reduction of time cost in comparison with the standard MFML approach.

\subsection{Ground State and Vertical Excitation Energies}
\begin{figure}[htb!]
    \centering
    \includegraphics[width=\linewidth]{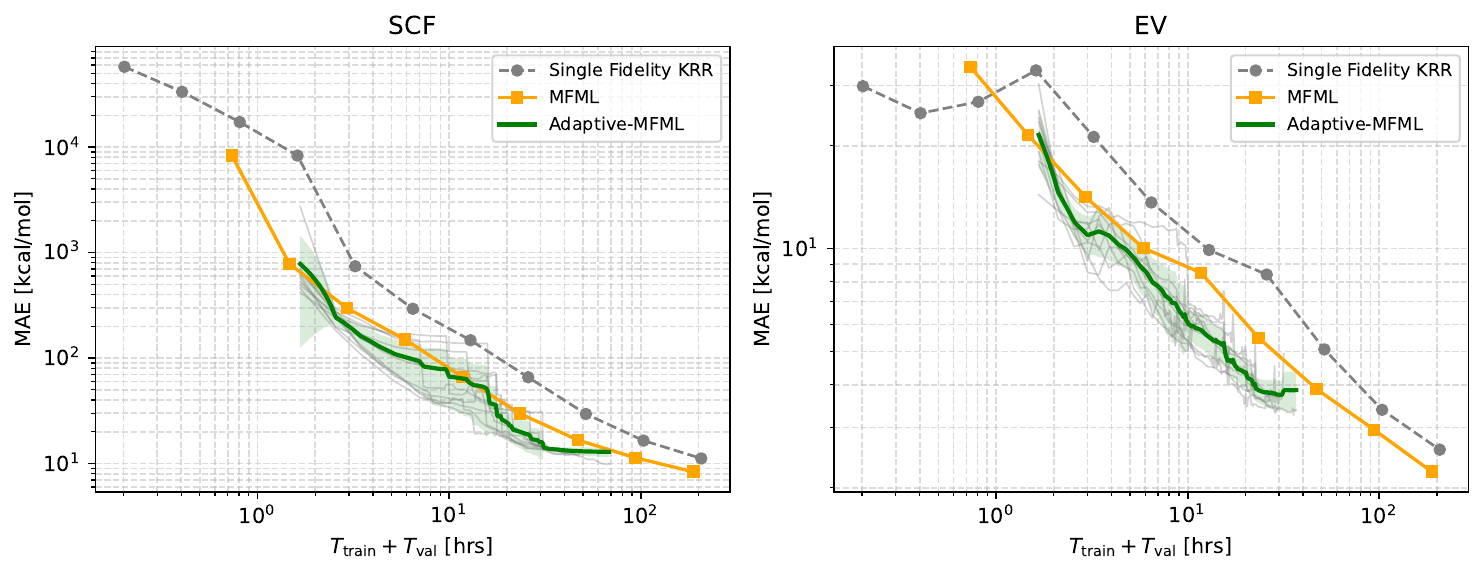}
    \caption{Model error (MAE) as a function of training data generation time-cost for single fidelity KRR, basic MFML, and adaptive-MFML on the QeMFi dataset \cite{vinod2024QeMFi_paper, vinod_2024_QeMFi_zenodo_datatset}. For a fixed time-cost budget the adaptive-MFML model gives a lower MAE than both single fidelity KRR and basic MFML. However, the stopping criterion causes it to terminate around 70~hrs.
}
    \label{fig_QeMFi_LC}
\end{figure}

The QeMFi dataset \cite{vinod2024QeMFi_paper, vinod_2024_QeMFi_zenodo_datatset} provides a more complex chemical challenge than the VIB5 dataset consisting of nine chemically diverse molecules with geometries sampled from Wigner distributions \cite{ws22_data_paper}. For each of the 135,000 geometries, electronic properties, including ground state self-consistent field (SCF) energies and vertical excitation (EV) energies, are computed using the TD-DFT formalism with the CAM-B3LYP functional. 
The multifidelity hierarchy is defined by increasing basis set sizes: 3-21G, 6-31G, def2-SVP, and the def2-TZVP target fidelity (see Supplementary Information Section S3 for details). The STO-3G level available in QeMFi is omitted due to our earlier work indicating issues with this level in MFML schemes \cite{vinod2024_gamma_curve_error_contours, vinod2024_nonnestedMFML}.

We evaluate the novel adaptive-MFML method on both the SCF energies and the more challenging vertical excitation energies. 
Fig.~\ref{fig_QeMFi_LC} presents the time-cost versus MAE curves for these properties. Analyzing the SCF energy predictions presents an alternative perspective on model performance, that is, evaluating accuracy under a strict computational budget as opposed to considering a target MAE (as studied in the case for VIB5). 
Given a fixed budget of, say, 30 hours, the single-fidelity KRR model reaches an MAE of $\sim$60~kcal/mol, the basic MFML model $\sim$25~kcal/mol, and the adaptive-MFML model $\sim$15~kcal/mol. This is nearly a 4-fold reduction relative to single-fidelity KRR and a 1.5$\times$ reduction relative to basic MFML. The advantage over basic MFML is therefore modest for the SCF energies, and it is bounded by the stopping rule. The adaptive run terminates at $\approx$13~kcal/mol after $\approx$70~h of total cost, while basic MFML shows improvement as
the budget grows to $\sim$200~hrs.

Considering the vertical excitation energies on the right-hand side of Fig.~\ref{fig_QeMFi_LC}, limiting the time-cost budget to about 20 hours results in an MAE of $\sim 9$ kcal/mol for the single fidelity KRR model and a $\sim6$ kcal/mol for the basic MFML model. The adaptive-MFML model reduces this error to $\sim4$ kcal/mol. While the relative performance gain of using the adaptive-MFML algorithm are markedly smaller here than those observed for VIB5, it is to be noted that this is due to the inherent complexity of the excitation state energy landscape which are known to be difficult for ML models to capture well \cite{Westermayr2020review, dral_molecular_2021}.
Despite this universal challenge across ML in QC, the adaptive-MFML algorithm outperforms both the single fidelity KRR and basic MFML baselines. 

\subsection{Coupled Cluster Energies of ANI-1ccx} \label{sec:ani_details}
\begin{figure}[htb!]
    \centering
    \includegraphics[width=0.65\linewidth]{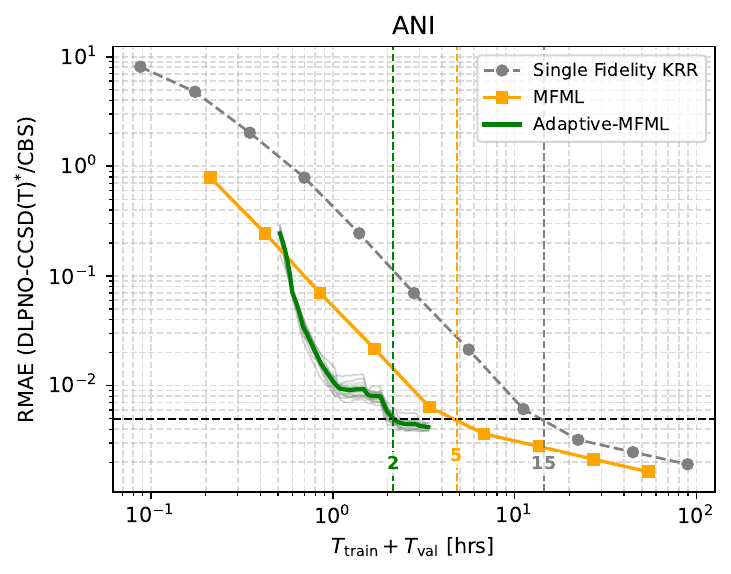}
    \caption{Training data generation time-cost versus RMAE for the ANI-1ccx dataset collection of molecules \cite{2020_Smith_ANI1ccx_dataset}. The adaptive-MFML method reaches a target error at the lowest overall time-cost.}
    \label{fig_ANI_LC}
\end{figure}

Coupled cluster (CC) methods are widely regarded as the golden standard in QC. However, computing high fidelity properties at CC becomes prohibitive for molecules with more than a dozen atoms \cite{Bartlett_CCSD_theory_2007}. 
The ANI-1ccx dataset consists of several molecules of varying sizes with the fidelities HF/ccpVDZ, DFT-$\omega$b97x/6-31G*, DFT-$\omega$b97x/def2-TZVPP, and the DLPNO-CCSD(T)*/CBS fidelities \cite{2020_Smith_ANI1ccx_dataset} (see Supplementary Information Section S3 for data pre-processing and Section S4 for time-cost computation protocols).
We contrast the performance of the single-fidelity KRR, basic MFML, and adaptive-MFML models for this hierarchy. 

Fig.~\ref{fig_ANI_LC} reports the learning curves for predicting molecular energies at the DLPNO-CCSD(T)*/CBS fidelity. 
A horizontal line is drawn at $5\times 10^{-3}$ for visually comparing the different methods. 
The single fidelity KRR model requires about 15~hrs of compute costs to reach the threshold. The basic MFML model takes only 5~hrs to hit that threshold. However, the adaptive-MFML model presented in this work takes as little as 2~hrs to achieve this threshold. This presents a $7.5\times$ improvement over single fidelity KRR and 2.5 fold improvement over basic MFML.
For larger cost budgets, the learning curves for all three models begin to plateau out and saturate. That is, additional training data does not yield as strong an improvement in the models' accuracy indicating an inherent structural complexity within the data. Crucially, since this saturation is observed for single fidelity KRR and the multifidelity models, it is not an artifact of the MFML or adaptive scheme itself but could be related to the intrinsic property of the chemical space being learned, or the KRR architecture, or the representation being used.

\begin{figure}[htb!]
    \centering
    \includegraphics[width=\linewidth]{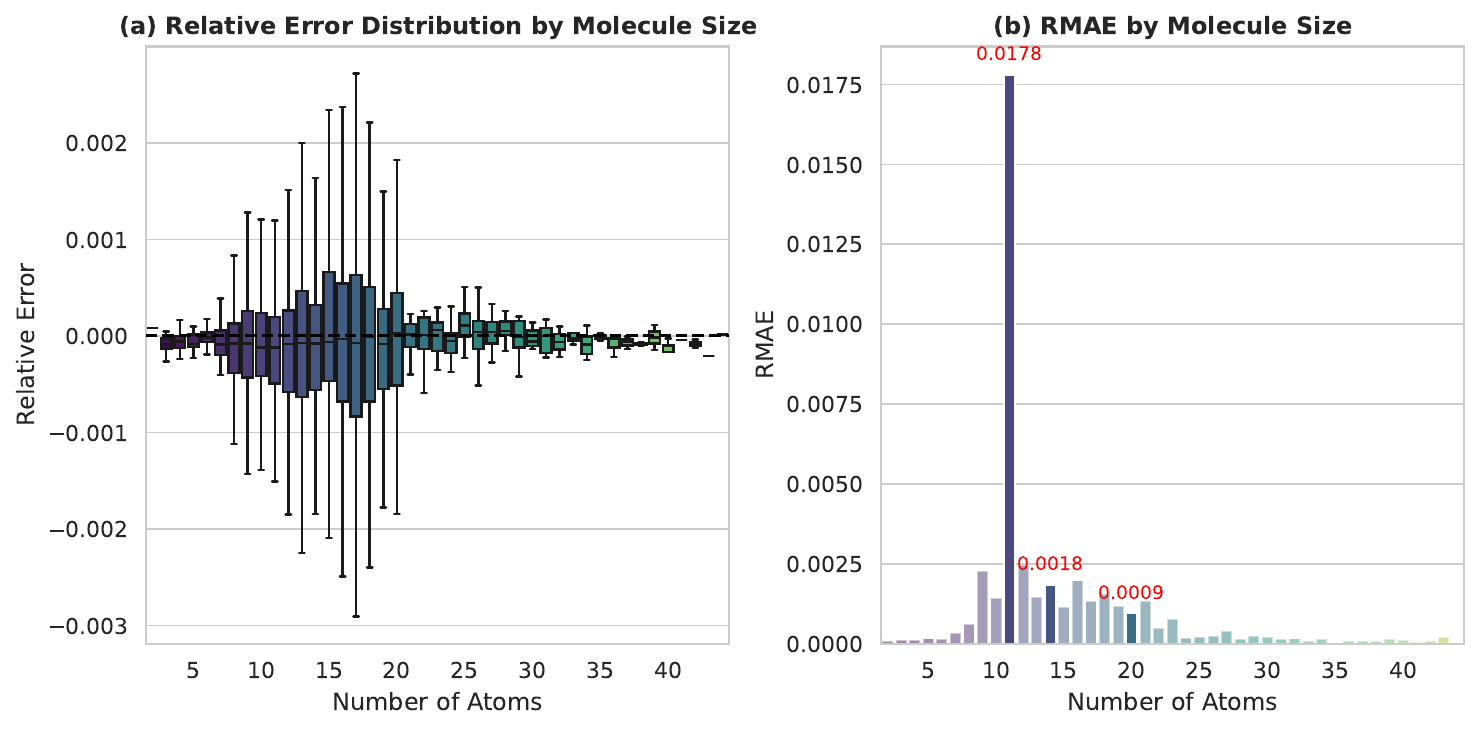}
    \caption{(a) Distribution of relative error for the adaptive-MFML model (trained for a $\sim3$hrs time-cost budget), categorized by molecule size in the ANI-1ccx dataset. (b) RMAE of the same adaptive-MFML model, binned by the number of atoms per molecule.}
    \label{fig_ANI_errordistr}
\end{figure}

To further analyze the adaptive-MFML model for the ANI-1ccx dataset, 
Fig.~\ref{fig_ANI_errordistr}(a) presents the distribution of differences between predictions and reference DLPNO-CCSD(T)*/CBS energies (that is $y^{F}-P_{\rm adaptive-MFML}$). We select the adaptive-MFML model trained under a $\sim3$ hour computational budget, which corresponds to the multifidelity training dataset composition of $[768,112,28,12]$ - which is the terminal location of the green curve from Fig.~\ref{fig_ANI_LC}. 
The differences are centered on zero RMAE indicating good agreement between prediction and reference in most cases. For molecules with number of atoms ranging between 8-25, the error bars extend to about 0.003 on either side of 0. While this might seem like a large fluctuation, it is worth noting that the energies themselves range between -400,000 kcal/mol and 200,000 kcal/mol as seen in Fig.~SF4 of the supplementary information. The relative error indicates at most a fluctuation of 0.3\% Indeed Fig.~SF4 presents that the predictions and references from this adaptive-MFML model lies close to the identity line indicating very good agreement. A detailed assessment on this can be found in section S7 of the supplementary information along with Table ST4 reporting the MAE, RMSE, and MAE per atom for the adaptive-MFML and single fidelity KRR model trained for a single seed.

Fig.~\ref{fig_ANI_errordistr}(b) presents RMAEs binned by the number of atoms in the molecules of the ANI-1ccx dataset. The highest recorded RMAE occurs in the 11-atom bin. Regardless, it is worth noting that this indicates that the prediction of the adaptive-MFML model is off by about 1.78\% on average from the true DLPNO-CCSD(T)*/CBS energy values.
Overall, the RMAE distribution on the basis of the number of atoms in the molecule have a low value for most of the bins underscoring the predictive accuracy of the adaptive-MFML model across varying molecular size. 

\begin{figure}[htb!]
    \centering
    \includegraphics[width=\linewidth]{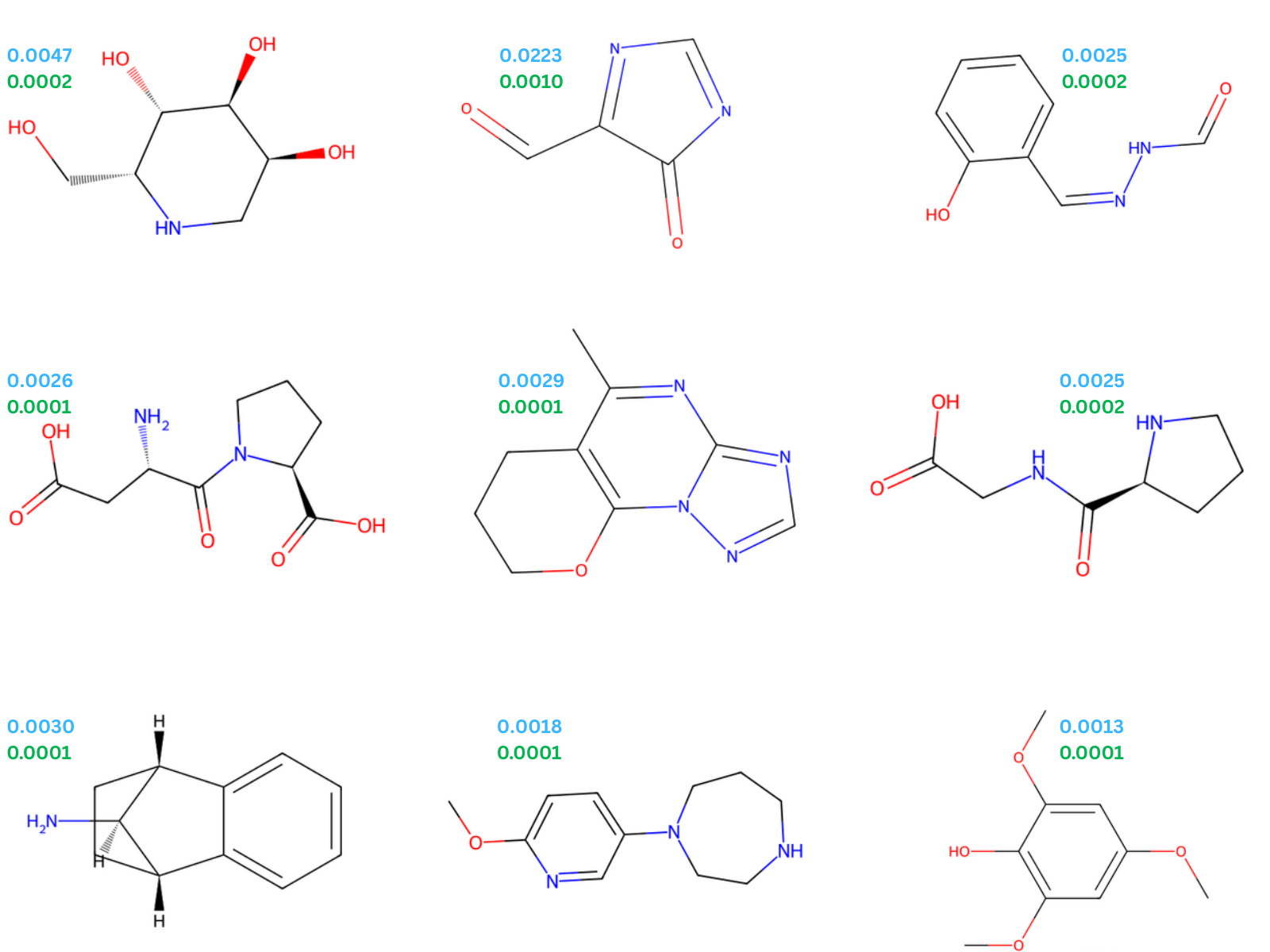}
    \caption{Relative error between predicted and reference DLPNO-CCSD(T)*/CBS energies for selected molecules for a single fidelity KRR model trained on $2^6$ DLPNO-CCSD(T)*/CBS samples ($\sim$3 hours of time-cost) - shown in blue text - and the adaptive-MFML model ($\sim 3$ hours of time-cost) - shown in green text. 
    The adaptive-MFML model consistently achieves the lowest relative error, confirming that the efficiency gain translates into robust individual molecular predictions.}
    \label{fig_ANI_molerrors}
\end{figure}
To substantiate the general statistical analysis above, we examine predictions for a few randomly selected molecules from the test set of the ANI-1ccx dataset. We compare two models: a single fidelity KRR model trained on $2^6$ DLPNO-CCSD(T)*/CBS data points (time-cost $\sim$3 hours), and the adaptive-MFML model with training dataset composition $[768,112,28,12]$ (time-cost $\sim$3 hours). The single fidelity KRR and adaptive-MFML models are trained with a similar time-cost budget. Fig.~\ref{fig_ANI_molerrors} reports the relative error of prediction, $\left\lvert y^F-P_{\rm ML}\right\rvert$, for a random selection of molecules from the test set. 
For all the molecules, the adaptive-MFML model consistently yields the lowest error. This confirms that the adaptive-MFML algorithm not only reduces training data costs to reach a target representative error, but that it also faithfully reproduces the high-fidelity reference energies for molecules of diverse sizes.

\section{Discussion}
The generation of high-fidelity training data remains a prohibitive hurdle for the successful and sustainable use of ML in QC. While standard MFML significantly reduces the computational cost, reliance on rigid heuristics for training dataset composition limits the maximum achievable efficiency in these methods. The adaptive algorithm for MFML that is developed in this work addresses this directly by dynamically growing the training datasets on-the-fly ensuring further streamlining of cost budgets in such models. Our numerical experiments validate the efficacy of this algorithm for diverse QC properties ranging from gold standard CCSD(T) energies to hard-to-learn excitation energies. While fixed scaling factor MFML assumes a uniform rate of information gain across all fidelities, the adaptive method inherently measures the marginal information gain of each batch of training samples added. If a lower fidelity does capture the underlying physical characteristic of the QC property, the algorithm `recognizes' this and limits the querying of high-fidelity reference computations.  
A variant of the herein developed adaptive algorithm is evaluated separately in Supplementary Information Section S8 to provide an exhaustive analysis of this approach for MFML.

While the adaptive-MFML method works across different QC properties as evidenced in the results above, it is also to be noted that for the ANI-1ccx dataset, the wall-time computations were measured using a static proxy based on the mean molecular size of 14 atoms, rather than exact per-molecule timings. Because our adaptive sampling process draws molecules randomly from a distribution ranging from 2 to 43 atoms, the realized computational cost is highly dependent on the specific draw. Consequently, the true cost of generating a given training composition carries an inherent variance. 
Furthermore, the efficiency of any multifidelity approach is fundamentally driven by the cost ratio between the fidelities. Since CCSD(T) and HF exhibit vastly different formal scaling behaviors with respect to system size ($\mathcal{O}(N^7)$ versus $\mathcal{O}(N^4)$), this cost ratio is certainly not stable across the ANI-1ccx dataset but rather grows with the number of atoms. While our use of a proxy cost provides a statistically representative baseline, it does not perfectly reflect the cost variance that might occur due to the selection of random molecules of differing sizes. It is to be noted that this is a limitation not of the adaptive-MFML method but rather of the dataset itself. Future dataset curation should explicitly include per-calculation wall-times, which would allow subsequent studies to perform rigorous cost-variance analyses and establish precise error bars for multifidelity training compositions.

Several open areas of research can be identified as extensions of this work. Formulating the dataset size at each fidelity as hyper-parameters to be optimized via a differentiable objective function is one such promising direction. Recent advances in active learning to guide geometry sampling for the chemical configuration space using low fidelity reference data promise improvement in prediction accuracy by an order of magnitude \cite{vinod_2026_lfab}. The modular nature of the adaptive algorithm enables seamless integration with such active learning schemes to further improve the cost efficiency of multifidelity methods. The adaptive-MFML method is certainly not without limitations. One immediate observation is that dataset generation for standard MFML allows for parallelization of training data generation across fidelities whereas in the adaptive-MFML algorithm, since reference calculations are on demand, it introduces a sequential limitation on the data generation. However, the efficacy of the method ought to make up for this limitation. One potential solution to this serialization is the use of pre-trained global machine learning interatomic potentials (MLIPs) \cite{Smith_Nebgen_Lubbers_Isayev_Roitberg_2018_anipotential, Chen21_CCSD_MLPotential} to further reduce the cost of making actual computations. Such an approach would come with its own challenge of recursive training of ML models on their own outputs and is an interesting area of research.
In our practice, even on CPU-based hardware (that is no parallelization or GPU architecture), the training of the MFML model with several fidelities takes at most a second. The prediction routine (with a small validation set for instance) is much less. Overall, the cost of training-prediction with the MFML models using KRR architecture has been negligible. As a concrete example, a single run of the adaptive-MFML algorithm for $\rm CH_3Cl$ of VIB5 took $\sim74.28$ seconds for the process to terminate. Of course, here the training data was pre-existing. This indicates that the predominant cost of running the adaptive scheme will come from generating the training data and not the method itself. That is, the cost of the algorithm itself is nearly negligible.

However, we also note that the adaptive-MFML scheme discussed in this work uses batch addition of training data per fidelity. This allows for parallelization of these computations. 
Furthermore, recent advances in pool-based active learning \cite{vinod_2026_lfab} would be a tool to first sort the data-pool into importance by geometry and thereafter run the adaptive-MFML scheme for a much larger batch size allowing for better parallel deployment in clusters. Additionally, one could treat the number of training samples at each fidelity to belong to an isotropic optimization problem accounting for both cost and accuracy offered by several small sized seed-MFML models. Alternative strategies could include using results from this work to notice that the adaptive-MFML model favors the lower fidelity (in spite of having no information about the cost per fidelity!) and setting the initial MFML model in the algorithm to take a very large number of training samples at the lowest fidelity thereby reducing the number of epochs that would occur.

Overall the adaptive-MFML method represents a substantial leap forward in cost-aware sustainable QC. It offers a noticeable improvement not only over single fidelity methods but also over conventional MFML paradigms. By dynamically mitigating data redundancy, it provides a highly efficient solution across diverse QC properties and complex molecular spaces. We have made the associated source code fully open-access and well documented, ensuring this adaptive framework is readily deployable for the broader computational community.

\backmatter

\section*{Data availability}
All data used in this work comes from openly available datasets which are appropriately cited.

\section*{Code Availability}
The programming scripts used for this work can be openly accessed at \href{https://github.com/SM4DA/Adaptive-MFML}{this GitHub repository}. 

\section*{Supplementary information}
Supplementary Sections S1--S10, Figures SF1--SF14, Tables ST1--ST6, Algorithm SA1.

\section*{Acknowledgments}
The authors acknowledge support by the DFG through the project ZA 1175/3-1 as well as through the DFG Priority Program SPP 2363 on `Utilization and Development of Machine Learning for Molecular Applications – Molecular Machine Learning' through the project ZA 1175/4-1. The authors would also like to acknowledge the support of the `Interdisciplinary Center for Machine Learning and Data Analytics (IZMD)' at the University of Wuppertal. VV thanks Sergei Tretiak, Olexandr Isayev, and Justin Smith for guidance with the ANI-1ccx dataset. We also thank the three anonymous reviewers for a very constructive criticism which improved the quality of this manuscript and its experiments significantly. VV also thanks Louise Schaub for her design feedback. 

\section*{Author Contribution}
VV: writing, revision, conceptualization, methodology, validation, formal analysis, investigation, data curation, visualization\\ 
PZ: conceptualization, validation, supervision, funding acquisition 

\section*{Declarations}
The authors declare that there is no conflict of interest or any competing interests. 

\bibliography{main.bib}

@article{Kennedy_2000_MF_fastapprox_copmutercodes, 
title={Predicting the output from a complex computer code when fast approximations are available}, 
volume={87}, 
ISSN={0006-3444}, 
DOI={10.1093/biomet/87.1.1}, 
number={1}, 
journal={Biometrika}, 
author={Kennedy, MC and O’Hagan, A}, 
year={2000}, 
month={03}, 
pages={1–13} 
}

@article{Fernandez-Godino_2023, 
title={Review of multi-fidelity models}, 
volume={1}, 
DOI={10.3934/acse.2023015},
URL={https:/doi.org/10.3934/acse.2023015},
number={4}, 
journal={Adv. Comput. Sci. Eng.}, 
publisher={Advances in Computational Science and Engineering}, 
author={Fernández-Godino, M. Giselle}, 
year={2023}, 
month={12}, 
pages={351–400} 
}

@book{Crawford_CCSD_theory_2000,
author = {Crawford, T. Daniel and Schaefer III, Henry F.},
publisher = {John Wiley \& Sons, Ltd},
isbn = {9780470125915},
title = {An Introduction to Coupled Cluster Theory for Computational Chemists},
booktitle = {Rev. Comput. Chem.},
chapter = {2},
pages = {33-136},
doi = {10.1002/9780470125915.ch2},
year = {2000},
address={~}
}

@article{vinod23_MFML,
author = {Vinod, Vivin and Maity, Sayan and Zaspel, Peter and Kleinekathöfer, Ulrich},
title = {Multifidelity Machine Learning for Molecular Excitation Energies},
journal = {J. Chem. Theory Comput.},
volume = {19},
number = {21},
pages = {7658-7670},
year = {2023},
doi = {10.1021/acs.jctc.3c00882}
}

@article{vinod_2024_oMFML,
doi = {10.1088/2632-2153/ad2cef},
year = {2024},
month = {03},
publisher = {IOP Publishing},
volume = {5},
number = {1},
pages = {015054},
author = {Vivin Vinod and Ulrich Kleinekathöfer and Peter Zaspel},
title = {Optimized multifidelity machine learning for quantum chemistry},
journal = {Mach. Learn.: Sci. Technol.}
}

@article{vinod2024_benchmarking_dataefficiency_deltaml,
      author = {Vinod, Vivin and Zaspel, Peter},
    title = {Benchmarking data efficiency in {$\Delta$-ML} and multifidelity models for quantum chemistry},
    journal = {J. Chem. Phys.},
    volume = {163},
    number = {2},
    pages = {024134},
    year = {2025},
    month = {07},
    issn = {0021-9606},
    doi = {10.1063/5.0272457},
}

@article{lyu2024_porphyrin,
      author = {Lyu, Dongyu and Vinod, Vivin and Holzenkamp, Matthias and Holtkamp, Yannick Marcel and Maity, Sayan and Salazar, Carlos R. and Kleinekathöfer, Ulrich and Zaspel, Peter},
title = {Excitation Energy Transfer between Porphyrin Dyes on a Clay Surface: A Study Employing Multifidelity Machine Learning},
journal = {Adv. Theory Simul.},
volume = {8},
number = {11},
pages = {e00271},
doi = {https://doi.org/10.1002/adts.202500271},
year = {2025}
}

@article{vinod2024_gamma_curve_error_contours,
author = {Vinod, Vivin and Zaspel, Peter},
title = {Investigating Data Hierarchies in Multifidelity Machine Learning for Excitation Energies},
journal = {J. Chem. Theory Comput.},
volume = {21},
number = {6},
pages = {3077-3091},
year = {2025},
doi = {10.1021/acs.jctc.4c01491}
}

@Article{zasp19a,
  author    = {Zaspel, Peter and Huang, Bing and Harbrecht, Helmut and Von Lilienfeld, O. Anatole},
  journal   = {J. Chem. Theory Comput.},
  title     = {Boosting Quantum Machine Learning Models with a Multilevel Combination Technique: {Pople Diagrams} Revisited},
  year      = {2019},
  number    = {3},
  pages     = {1546--1559},
  volume    = {15},
  doi       = {10.1021/acs.jctc.8b00832},
  publisher = {American Chemical Society},
}

@article{dral2020hierarchical,
  title={Hierarchical machine learning of potential energy surfaces},
  author={Dral, Pavlo O and Owens, Alec and Dral, Alexey and Cs{\'a}nyi, G{\'a}bor},
  journal={J. Chem. Phys.},
  volume={152},
  number={20},
  doi = {10.1063/5.0006498},
  pages={204110},
  year={2020},
  publisher={AIP Publishing LLC}
}

@article{zhang_vib5_2022,
    title = {{VIB5} database with accurate ab initio quantum chemical molecular potential energy surfaces},
    volume = {9},
    copyright = {2022 The Author(s)},
    issn = {2052-4463},
    doi = {10.1038/s41597-022-01185-w},
    number = {1},
    journal = {Sci. Data},
    author = {Zhang, Lina and Zhang, Shuang and Owens, Alec and Yurchenko, Sergei N. and Dral, Pavlo O.},
    month = {03},
    year = {2022},
    pages = {84}
}

@Article{vinod_2024_QeMFi_zenodo_datatset,
  author       = {Vinod, Vivin and Zaspel, Peter},
  title        = {{QeMFi}: A Multifidelity Dataset of Quantum Chemical 
                   Properties of Diverse Molecules (1.1.0) [dataset]},
  month        = {10},
  year         = 2024,
  publisher    = {Zenodo},
  version      = {1.1.0},
  doi          = {10.5281/zenodo.13925688},
journal = {Zenodo}
}

@article{vinod2024QeMFi_paper, 
title={{QeMFi}: A Multifidelity Dataset of Quantum Chemical Properties of Diverse Molecules}, 
volume={12}, 
ISSN={2052-4463}, 
DOI={10.1038/s41597-024-04247-3}, 
number={1}, 
journal={Sci. Data},
author = {Vinod, Vivin and Zaspel, Peter}, 
year={2025}, 
month={02}, 
pages={202}
}

@article{vinod2024_nonnestedMFML,
    title={{Assessing} non-nested configurations of multifidelity machine learning for quantum-chemical properties}, 
    volume={5}, 
    ISSN={2632-2153}, 
    DOI={10.1088/2632-2153/ad7f25}, 
    number={4}, 
    journal={Mach. Learn.: Sci. Technol.}, 
    publisher={IOP Publishing}, 
    author={Vinod, Vivin and Zaspel, Peter}, 
    year={2024}, 
    month={10}, 
    pages={045005}
}

@article{vinod_2026_lfab,
author = {Vinod, Vivin and Zaspel, Peter},
title = {{LFaB:} Low fidelity as Bias for Active Learning in the chemical configuration space},
journal = {J. Chem. Theory Comput.},
volume = {22},
number = {11},
pages = {5637-5648},
year = {2026},
doi = {10.1021/acs.jctc.6c00009},
}

@article{ws22_data_paper,
author={Pinheiro Jr, Max
and Zhang, Shuang
and Dral, Pavlo O.
and Barbatti, Mario},
title={{WS22 database, Wigner Sampling} and geometry interpolation for configurationally diverse molecular datasets},
journal={Sci. Data},
year={2023},
month={02},
day={15},
volume={10},
number={1},
pages={95},
issn={2052-4463},
doi={10.1038/s41597-023-01998-3}
}

@Article{Westermayr2020review,
  author    = {Julia Westermayr and Philipp Marquetand},
  journal   = {Chem. Rev.},
  title     = {Machine Learning for Electronically Excited States of Molecules},
  year      = {2020},
  month     = {11},
  number    = {16},
  pages     = {9873--9926},
  volume    = {121},
  doi       = {10.1021/acs.chemrev.0c00749},
  publisher = {American Chemical Society ({ACS})},
}

@article{dral_molecular_2021,
    title = {Molecular excited states through a machine learning lens},
    volume = {5},
    copyright = {2021 Springer Nature Limited},
    issn = {2397-3358},
    doi = {10.1038/s41570-021-00278-1},
    language = {en},
    number = {6},
    journal = {Nat. Rev. Chem.},
    author = {Dral, Pavlo O. and Barbatti, Mario},
    month = {06},
    year = {2021},
    pages = {388--405},
}

@article{2020_Smith_ANI1ccx_dataset, 
title={The {ANI-1ccx and ANI-1x} data sets, coupled-cluster and density functional theory properties for molecules}, 
volume={7}, 
ISSN={2052-4463}, 
DOI={10.1038/s41597-020-0473-z}, 
number={1}, 
journal={Sci. Data}, 
publisher={Nature Publishing Group}, 
author={Smith, Justin S. and Zubatyuk, Roman and Nebgen, Benjamin and Lubbers, Nicholas and Barros, Kipton and Roitberg, Adrian E. and Isayev, Olexandr and Tretiak, Sergei}, 
year={2020}, 
month={05}, 
pages={134} 
}

@Article{Pilania2017,
  author   = {Pilania, G. and Gubernatis, J. E. and Lookman, T.},
  journal  = {Comput. Mater. Sci.},
  title    = {Multi-fidelity machine learning models for accurate bandgap predictions of solids},
  year     = {2017},
  issn     = {0927-0256},
  month    = {03},
  pages    = {156--163},
  volume   = {129},
  doi      = {10.1016/j.commatsci.2016.12.004},
}

@article{ANI-NN_ML,
author = {Gao, Xiang and Ramezanghorbani, Farhad and Isayev, Olexandr and Smith, Justin S. and Roitberg, Adrian E.},
title = {{TorchANI}: A Free and Open Source {PyTorch}-Based Deep Learning Implementation of the ANI Neural Network Potentials},
journal = {J. Chem. Inf. Modeling},
volume = {60},
number = {7},
pages = {3408-3415},
year = {2020},
doi = {10.1021/acs.jcim.0c00451}
}

@article{Bungartz_Griebel_2004, title={Sparse grids}, volume={13}, DOI={10.1017/S0962492904000182}, journal={Acta Numerica}, author={Bungartz, Hans-Joachim and Griebel, Michael}, year={2004}, pages={147–269}}

@Article{smit17_ANI-1,
  author    = {Smith, J. S. and Isayev, O. and Roitberg, A. E.},
  journal   = {Chem. Sci.},
  title     = {{ANI-1}: An Extensible Neural Network Potential with {DFT} Accuracy at Force Field Computational Cost},
  year      = {2017},
  issn      = {2041-6520},
  number    = {4},
  pages     = {3192--3203},
  volume    = {8},
  doi       = {10.1039/C6SC05720A},
  publisher = {The Royal Society of Chemistry},
}

@article{lilienfeld_2015_descriptors_requirement,
author = {von Lilienfeld, O. Anatole and Ramakrishnan, Raghunathan and Rupp, Matthias and Knoll, Aaron},
title = {{Fourier} series of atomic radial distribution functions: A molecular fingerprint for machine learning models of quantum chemical properties},
journal = {Int. J. Quantum Chem.},
volume = {115},
number = {16},
pages = {1084-1093},
doi = {10.1002/qua.24912},
year = {2015}
}

@article{RupCM,
  title = {Fast and Accurate Modeling of Molecular Atomization Energies with Machine Learning},
  author = {Rupp, Matthias and Tkatchenko, Alexandre and M\"uller, Klaus-Robert and von Lilienfeld, O. Anatole},
  journal = {Phys. Rev. Lett.},
  volume = {108},
  issue = {5},
  pages = {05830-1 - 05830-5},
  year = {2012},
  month = {Jan},
  publisher = {American Physical Society},
  doi = {10.1103/PhysRevLett.108.058301}
}

@article{Vinod2024_multifidelity_monomers_CCSDT,
  author = {Vinod, Vivin and Lyu, Dongyu and Ruth, Marcel and R. Schreiner, Peter and Kleinekathöfer, Ulrich and Zaspel, Peter},
title = {Predicting Molecular Energies of Small Organic Molecules With Multi-Fidelity Methods},
journal = {J. Comp. Chem.},
volume = {46},
number = {6},
pages = {e70056},
doi = {https://doi.org/10.1002/jcc.70056},
year = {2025}
}

@article{Huang2020slatm,
  title = {Quantum machine learning using atom-in-molecule-based fragments selected on the fly},
  author = {Huang, Bing and von Lilienfeld, O Anatole},
  journal = {Nat. Chem.},
  volume =  {12},
  number =  {10},
  pages = {945--951},
  month =  {10},
  year =  {2020},
  doi = {10.1038/s41557-020-0527-z}
}

@article{dral2020quantum,
  title={Quantum chemistry in the age of machine learning},
  doi = {10.1021/acs.jpclett.9b03664},
  author={Pavlo O Dral},
  journal={J. Phys. Chem. Lett.},
  volume={11},
  number={6},
  pages={2336--2347},
  year={2020},
  publisher={ACS Publications}
}

@article{westermayr_2021_perspective,
    author = {Westermayr, Julia and Gastegger, Michael and Schütt, Kristof T. and Maurer, Reinhard J.},
    title = "{Perspective on integrating machine learning into computational chemistry and materials science}",
    journal = {J. of Chem. Phys.},
    volume = {154},
    number = {23},
    pages = {230903},
    year = {2021},
    month = {06},
    issn = {0021-9606},
    doi = {10.1063/5.0047760}
}

@Article{Ramakrishnan2015,
  author     = {Ramakrishnan, Raghunathan and Dral, Pavlo O. and Rupp, Matthias and von Lilienfeld, O. Anatole},
  journal    = {J. Chem. Theory Comput.},
  title      = {Big Data Meets Quantum Chemistry Approximations: The {$\Delta$}-Machine Learning Approach},
  year       = {2015},
  issn       = {1549-9618},
  month      = {05},
  number     = {5},
  pages      = {2087--2096},
  volume     = {11},
  doi        = {10.1021/acs.jctc.5b00099},
  publisher  = {American Chemical Society}
}

@article{Ruth_delML_CCSD_DFT,
author = {Ruth, Marcel and Gerbig, Dennis and Schreiner, Peter R.},
title = {Machine Learning for Bridging the Gap between Density Functional Theory and Coupled Cluster Energies},
journal = {J. Chem. Theory and Comp.},
volume = {19},
number = {15},
pages = {4912-4920},
year = {2023},
doi = {10.1021/acs.jctc.3c00274}
}

@misc{sandonas2026perspectivesustainableexplorationchemical,
      title={Perspective: Towards sustainable exploration of chemical spaces with machine learning}, 
      author={Leonardo Medrano Sandonas and David Balcells and Anton Bochkarev and Jacqueline M. Cole and Volker L. Deringer and Werner Dobrautz and Adrian Ehrenhofer and Thorben Frank and Pascal Friederich and Rico Friedrich and Janine George and Luca Ghiringhelli and Alejandra Hinostroza Caldas and Veronika Juraskova and Hannes Kneiding and Yury Lysogorskiy and Johannes T. Margraf and Hanna Türk and Anatole von Lilienfeld and Milica Todorović and Alexandre Tkatchenko and Mariana Rossi and Gianaurelio Cuniberti},
      year={2026},
      eprint={2604.00069},
      archivePrefix={arXiv},
      primaryClass={cs.LG},
      url={https://arxiv.org/abs/2604.00069}, 
}

@article{
vinod_2026_deltaBCL,
author = {Sayan Maity and Vivin Vinod and Peter Zaspel and Ulrich Kleinekathöfer},
title = {{$\Delta$}-Machine Learning for {LC-DFT}-level Excitation Energies of Bacteriochlorophyll Molecules in a {LH2} Complex},
journal = {ChemRxiv},
volume = {2026},
number = {0504},
pages = {},
year = {2026},
doi = {10.26434/chemrxiv.15002714/v1}
}

@article{Bartlett_CCSD_theory_2007,
  title = {Coupled-cluster theory in quantum chemistry},
  author = {Bartlett, Rodney J. and Musia\l{}, Monika},
  journal = {Rev. Mod. Phys.},
  volume = {79},
  issue = {1},
  pages = {291--352},
  numpages = {0},
  year = {2007},
  month = {2},
  publisher = {American Physical Society},
  doi = {10.1103/RevModPhys.79.291}
}

@article{Smith_Nebgen_Lubbers_Isayev_Roitberg_2018_anipotential, 
title={{Less is more: Sampling chemical space with active learning}}, 
volume={148}, 
ISSN={0021-9606}, 
DOI={10.1063/1.5023802}, 
number={24}, 
journal={J. Chem. Phys.}, 
author={Smith, Justin S. and Nebgen, Ben and Lubbers, Nicholas and Isayev, Olexandr and Roitberg, Adrian E.}, 
year={2018}, 
month={05}, 
}

@article{Chen21_CCSD_MLPotential,
author = {Qu, Chen and Houston, Paul L. and Conte, Riccardo and Nandi, Apurba and Bowman, Joel M.},
title = {Breaking the Coupled Cluster Barrier for Machine-Learned Potentials of Large Molecules: The Case of 15-Atom Acetylacetone},
journal = {J. Phys. Chem. Lett.},
volume = {12},
number = {20},
pages = {4902-4909},
year = {2021},
doi = {10.1021/acs.jpclett.1c01142}
}

@article{Gratiet_Garnier_2014, 
title={RECURSIVE CO-KRIGING MODEL FOR DESIGN OF COMPUTER EXPERIMENTS WITH MULTIPLE LEVELS OF FIDELITY}, 
volume={4}, 
DOI={10.1615/Int.J.UncertaintyQuantification.2014006914}, 
number={5}, 
journal={Int. J. Uncertainty Quantif.}, 
author={Gratiet, Loic Le and Garnier, Josselin}, 
year={2014}}

@article{Heinen_Khan_Falk_Lilienfeld_2024, 
 title={Reducing training data needs with minimal multilevel machine learning {(M3L)}}, 
 volume={5}, ISSN={2632-2153}, DOI={10.1088/2632-2153/ad4ae5}, number={2}, journal={Mach. Learn.: Sci. Technol.}, author={Heinen, Stefan and Khan, Danish and Falk von Rudorff, Guido and Karandashev, Konstantin and Jose Arismendi Arrieta, Daniel and Price, Alastair J A and Nandi, Surajit and Bhowmik, Arghya and Hermansson, Kersti and Anatole von Lilienfeld, O}, year={2024}, month={06}, pages={025058}}

@InProceedings{pmlr_v89_song19b,
  title = 	 {A General Framework for Multi-fidelity {Bayesian} Optimization with {Gaussian Processes}},
  author =       {Song, Jialin and Chen, Yuxin and Yue, Yisong},
  booktitle = 	 {Proceedings of the Twenty-Second International Conference on Artificial Intelligence and Statistics},
  pages = 	 {3158--3167},
  year = 	 {2019},
  editor = 	 {Chaudhuri, Kamalika and Sugiyama, Masashi},
  volume = 	 {89},
  series = 	 {Proceedings of Machine Learning Research},
  month = 	 {04},
  publisher =    {PMLR},
  url = 	 {https://proceedings.mlr.press/v89/song19b.html}
}

@InProceedings{pmlr_v151_li22b,
  title = 	 {Deep Multi-Fidelity Active Learning of High-Dimensional Outputs },
  author =       {Li, Shibo and Wang, Zheng and Kirby, Robert and Zhe, Shandian},
  booktitle = 	 {Proceedings of The 25th International Conference on Artificial Intelligence and Statistics},
  pages = 	 {1694--1711},
  year = 	 {2022},
  editor = 	 {Camps-Valls, Gustau and Ruiz, Francisco J. R. and Valera, Isabel},
  volume = 	 {151},
  series = 	 {Proceedings of Machine Learning Research},
  month = 	 {03},
  publisher =    {PMLR},
  url = 	 {https://proceedings.mlr.press/v151/li22b.html}
}

@article{Phan_Byun_2025, title={Cluster-aware threshold optimization with active learning for multi-fidelity surrogates},  url={https://doi.org/10.1007/s11081-025-10009-w}, DOI={10.1007/s11081-025-10009-w}, journal={Optim. Eng.}, author={Phan, Mai Thi Hong and Byun, Jai-Hyun}, year={2025}, month={08}}

\clearpage
\setcounter{section}{0}
\setcounter{figure}{0}
\setcounter{table}{0}
\setcounter{equation}{0}
\setcounter{algorithm}{0}
\setcounter{page}{1}
\renewcommand{\theequation}{SE\arabic{equation}}
\renewcommand{\thefigure}{SF\arabic{figure}}
\renewcommand{\thepage}{S\arabic{page}}
\renewcommand{\thesection}{S\arabic{section}}
\renewcommand{\thetable}{ST\arabic{table}}
\renewcommand{\thealgorithm}{SA\arabic{algorithm}}

\section*{Supplementary Information: Improvise, Adapt, Overcome --- An On-The-Fly Multifidelity Algorithm for Efficient Machine Learning}
\markboth{Supplementary Information}{Supplementary Information}

\newpage
\section{Kernel Ridge Regression} \label{sec:KRR}
Consider a training dataset at a fidelity $f$ defined as $\mathcal{T}^f:\{(\boldsymbol{x}_i,y_i^f)\}_{i=1}^{N_f}$. A kernel ridge regression (KRR) model approximates the prediction for an unseen query input $\boldsymbol{x}_q$ as 
\begin{equation}
    P_{\rm KRR}^f(\boldsymbol{x}_q):= \sum_{i=1}^{N_f}\alpha^f_ik(\boldsymbol{x}_1,\boldsymbol{x}_q), 
    \label{eq_KRR}
\end{equation}
where $k(\cdot,\cdot):\mathbb{R}^D\times\mathbb{R}^D\rightarrow\mathbb{R}$ is a kernel function. The coefficients $\alpha^f_i$ are computed by solving the linear system of equations 
\[
\left(\boldsymbol{K}+\lambda\boldsymbol{I}\right)\boldsymbol{\alpha} = \boldsymbol{Y}~.
\]
Here, $\boldsymbol{K}_{i,j}=k(\boldsymbol{x}_i,\boldsymbol{x}_j)$ is the kernel matrix for all pairs of inputs from $\mathcal{T}^f$, $\lambda$ is the Lavrentiev regularization term, $\boldsymbol{I}$ is the identity matrix, $\boldsymbol{\alpha}=[\alpha_i^f,\alpha_2^f,\ldots,\alpha_{N_f}^f]^{\top}$, and $\boldsymbol{Y}:=[y_1^f,y_2^f,\ldots,y_{N_f}^f]^{\top}$. 

The kernel function used in this work is the Matern kernel of second order with the $L_2$ distance metric. This is computed for each $(\boldsymbol{x},\boldsymbol{x}')$ pair as:
\begin{equation}
    k(\boldsymbol{x},\boldsymbol{x}') = \exp{\left(-\frac{\sqrt{3}}{\sigma}\left\lVert \boldsymbol{x}-\boldsymbol{x}'\right\rVert_2^2\right)}\cdot\left(1+\frac{\sqrt{3}}{\sigma}\left\lVert \boldsymbol{x}-\boldsymbol{x}'\right\rVert_2^2\right)~,
    \label{eq_matern}
\end{equation}
where $\sigma$ is a length-scale parameter for the kernel. 

\section{Hyperparameter Optimization}
\begin{table}[htb!]
    \centering
    \begin{tabular}{|c|c|c|}
         \hline
         Dataset (molecule/property) & Kernel Width ($\sigma$) & Regularization ($\lambda$) \\
         \hline
         \hline
         VIB5 ($\rm CH_3Cl$) & 170&$6\cdot10^{-10}$\\
         VIB5 ($\rm CH_3F$) & 50&$5\cdot10^{-11}$\\
         QeMFi (SCF) & 70&$6\cdot10^{-10}$\\
         QeMFi (EV) & 1600&$1\cdot10^{-14}$\\
         ANI & 80000&$2\cdot10^{-12}$\\
         \hline
    \end{tabular}
    \caption{Kernel hyperparameters used for different datasets in the main work.}
    \label{tab:hyperparams}
\end{table}
In the KRR model used in this work, there are two main hyper-parameters that are to be identified. This is the regularization $\lambda$ and kernel width $\sigma$. 
Identifying the hyperparameters is crucial to the ML pipeline employed. For the different datasets, the hyperparameters were optimized on a small portion of the training dataset with the highest fidelity of the dataset. The optimization was carried out using a the \texttt{gp\_minimize} package from the \texttt{skopt} library. The objective function minimizes the MAE on a small portion of the training dataset at the highest fidelity. Instead of the time-intensive grid-search method, this work employs Gaussian Processes with Bayesian optimization. The general \textit{ansatz} is to approximate the function using GPs. The different hyper-parameters are assumed to follow a multivariate Gaussian distribution. The GP kernel between the parameters provides the covariance of the function. The choice of the next collection of parameters that are to be evaluated for the optimization problem is on the basis of the acquisition function over the Gaussian prior. Further details of the scheme can be found at \href{https://scikit-optimize.github.io/stable/modules/generated/skopt.gp_minimize.html}{this URL}. The resulting hyper-parameters for each dataset are reported in Table \ref{tab:hyperparams}. In this current work, since we use pre-computed datasets, the hyper-parameters could be determined \textit{a priori}. In practice, however, one does not have this liberty. Instead, the hyper-parameters can be tuned on-the-fly similar to the workflow in AL schemes or other related adaptive protocols where after a certain number of epochs have passed, one reruns the hyper-parameter optimization.

\section{Datasets}
The work in the main manuscript benchmarks the new adaptive algorithms for MFML on three datasets. Here we provide salient features of the used datasets. 

\subsection{VIB5}
The VIB5 database consists of high accuracy \textit{ab initio} potential energy surfaces (PES) for $\rm CH_3Cl$ and $\rm CH_3F$ \cite{zhang_vib5_2022}. For both molecules, \textit{ab initio} potential energies are computed with varying fidelities. In this work, the assumed hierarchy of fidelities in increasing order is MP2/aug-cc-pVQZ, CCSD(T)-F12b/cc-pVTZ-F12, CCSD(T)-F12b/cc-pVQZ-F12, and a Theoretical best estimate (TBE) along the same lines of ref.~\cite{dral2020hierarchical}.
The computational cost of each fidelities is recorded in Table \ref{tab:vib5_time_cost} and is taken from ref.~\cite{dral2020hierarchical}.
\begin{table}[htb!]
\centering
\caption{Computational time cost per fidelity level for the VIB5 dataset.}
\label{tab:vib5_time_cost}
\begin{tabular}{|l|c|}
\hline
\textbf{Fidelity Level} & \textbf{Time Cost (s)} \\ 
\hline
MP2/aug-cc-pVQZ & 296.85 \\
CCSD(T)-F12b/cc-pVTZ-F12 & 298.24 \\
CCSD(T)-F12b/cc-pVQZ-F12 & 2224.71 \\
Theoretical Best Estimate (TBE) & 96144.24 \\ 
\hline
\end{tabular}
\end{table}

\subsection{QeMFi}
The quantum chemistry multifidelity (QeMFi) dataset contains nine chemically diverse molecules DFT computed QC properties \cite{vinod2024QeMFi_paper, vinod_2024_QeMFi_zenodo_datatset}. These molecules include 2-(methyliminomethyl)phenol (SMA), 4-(dimethylamino) benzonitrile (DMABN), and 4-(2-hydroxybenzylidene)-1,2-dimethyl-1H-imidazol-5(4H)-one (o-HBDI). 
Excitation and ground state (SCF) energies are computed for each of geometry using the DFT formalism with the CAM-B3LYP functional. The fidelities are decided by the basis set size and are (in increasing order of size): STO-3G, 3-21G, 6-31G, def2-SVP, and def2-TZVP. The energies are computed for $9\times 15,000$ geometries. 
In this work the four-level hierarchy begins at 3-21G and targets def2-TZVP. The STO-3G level of the original dataset is omitted since it was identified in previous work with QeMFi as a fidelity with poor characteristics \cite{vinod2024_nonnestedMFML, vinod2024_gamma_curve_error_contours}. All nine molecules are pooled into a single prediction problem for the two properties studied here.

\subsection{ANI-1ccx}
\begin{figure}[htb!]
    \centering
    \includegraphics[width=0.65\linewidth]{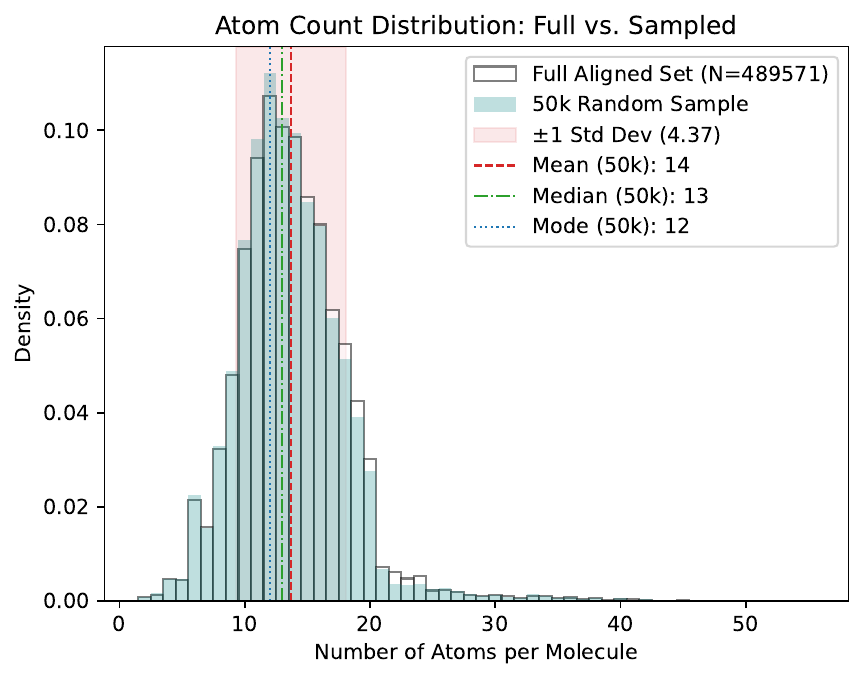}
    \caption{Distribution of number of atoms in molecules of the full ANI-1ccx dataset and the selected 50k samples for this work. There is good agreement between the two cases. The mean, median, mode, and standard deviation of number of samples is also shown.}
    \label{fig_ANI_hist}
\end{figure}
The ANI-1ccx dataset is a collection of molecular conformations with H, C, N, and O atoms. The original full database consists of roughly 500k unique geometries with corresponding energies computed with high accuracy DLPNO-CCSD(T)*/CBS level of theory \cite{ smit17_ANI-1, 2020_Smith_ANI1ccx_dataset, ANI-NN_ML}. For each of the geometries, several other fidelities are also available. In this work, we restrict ourselves to HF/ccpVDZ, DFT-$\omega$b97x/6-31G*, DFT-$\omega$b97x/def2-TZVPP, and the DLPNO-CCSD(T)*/CBS fidelities. To make the benchmarking of the adaptive sampling for MFML more tractable, we randomly sample 50k geometries from the parent dataset to use as the starting point. From this, we carry out a train:validation:test split of 85:5:10 resulting in 5,000 samples as the holdout test set. 

To verify that the sub-sampling carried out above did not disrupt any statistical properties of the dataset, we compare in Fig.~\ref{fig_ANI_hist}, the distribution of number of atoms in both the full ANI-1ccx and the sampled 50k geometries. There is good agreement between the two cases.

\section{Time-Cost Computation For ANI-1ccx}
\begin{figure}[htb!]
    \centering
    \includegraphics[width=0.65\linewidth]{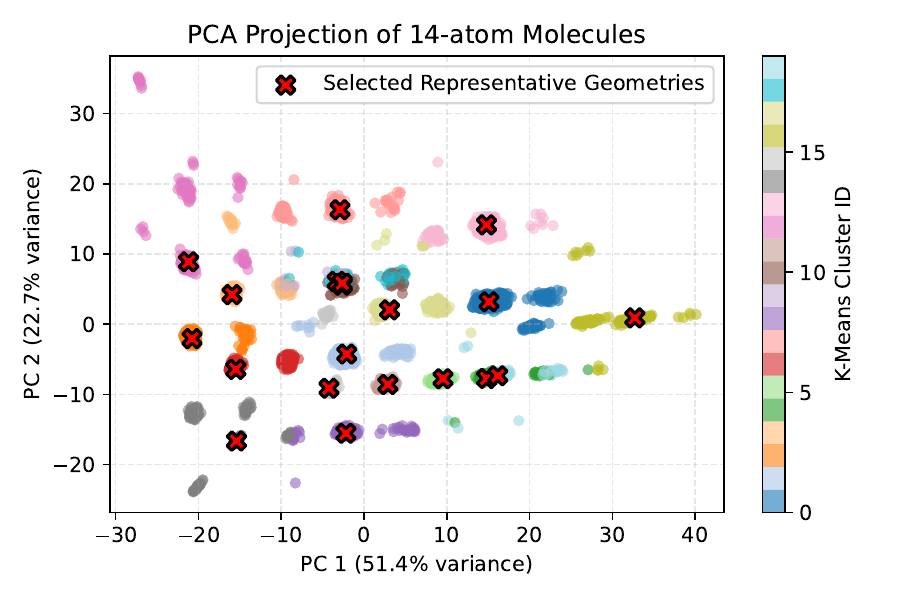}
    \caption{PCA decomposition of 14-atom molecules from the 50k subset of ANI-1ccx dataset. The geometries that are selected to compute time-costs are indicated with a red cross.}
    \label{fig_ANI_clustering}
\end{figure}
The ANI datasets do not provide time-costs of each computation. This makes them unsuitable to duly test the effectiveness of the multifidelity scheme in terms of reducing training data costs. In order to make the ANI dataset more accessible to the community to benchmark models with respect to training data cost and not simply model error, we carry out a brief strategy to compute time costs. First, having identified that the mean number of atoms in the ANI-1ccx dataset is 14 as seen in Fig.~\ref{fig_ANI_hist}, we isolate all molecules from the 50k dataset with 14 atoms. Next, a k-mean clustering is carried out for the molecular descriptors of these geometries with $k=20$ to select 20 geometries. Visually, this can be seen in Fig.~\ref{fig_ANI_clustering} where we have used PCA to project the clusters into the 2-D space.
For these selected geometries, we run ORCA simulations (see associated GitHub repository for reproducible scripts) computing the energies and recording the time taken to compute the energies. The computed time-costs are reported in Table \ref{tab:ani_time_cost}. It is to be noted, that the use of 20 molecules of 14-atom sizes is only statistically indicative of the cost incurred in the computations of the ANI-1ccx dataset. The diverse sizes of the molecules ranging from 2-43 provides a significant difference in the actual costs that may be incurred. This further highlights the pertinence of including time-costs with benchmark datasets for better evaluation of novel data-efficient schemes. 

While computing the DFT energies and their time cost is merely running the single ORCA script for them, it is slightly more nuanced for the DLPNO-CCSD(T)*/CBS dataset. The CBS energy is a composite of several fragments of energies \cite{2020_Smith_ANI1ccx_dataset}:
\begin{align}
E_{\mathrm{CBS}}^{\mathrm{CCSD(T)*}} &= E_{\mathrm{xtrap}}(3/4, \mathrm{HF}) + E_{\mathrm{xtrap}}(3/4, \mathrm{MP2}) - E(3, \mathrm{MP2}) \nonumber \\
& + E(3, \mathrm{NPNO-CCSD(T)}) + E(2, \mathrm{TPNO-CCSD(T)}) \nonumber \\
& - E(2, \mathrm{NPNO-CCSD(T)})~,\label{eq_CBS_Extrap}
\end{align}
where \(E(N, \mathrm{Method})\) 
is the resulting energy of `Method' computed with basis set of cardinal number $N\in\{2,3,4\}$ which stand for cc-pVDZ, cc-pVTZ, and cc-pVQZ respectively. NPNO and TPNO are Normal and Tight DLPNO respectively. 
Furthermore, \(\mathrm{Extrap}(N1/N2, \mathrm{Method})\) refers to the complete basis set extrapolation. It is also that
\begin{equation}
E_{\mathrm{xtrap}}(3/4, \mathrm{HF}) = \frac{e^{-\alpha\sqrt{4}}E_{\mathrm{HF}}^{(3)} - e^{-\alpha\sqrt{3}}E_{\mathrm{HF}}^{(4)}}{e^{-\alpha\sqrt{4}} - e^{-\alpha\sqrt{3}}}~,
\end{equation}
and
\begin{equation}
E_{\mathrm{xtrap}}(3/4, \mathrm{corr}) = \frac{4^{\beta} E_{\mathrm{corr}}^{(4)} - 3^{\beta} E_{\mathrm{corr}}^{(3)}}{4^{\beta} - 3^{\beta}}~.
\end{equation}
Thus a calculation of the DLPNO-CCSD(T)*/CBS needs five separate single‑point calculations, each supplying one or more of the terms in Eq.~\eqref{eq_CBS_Extrap}:
\begin{enumerate}
  \item DLPNO-CCSD(T)/cc-pVDZ, NormalPNO - supplies
        \(E(2,\mathrm{NPNO\text{-}CCSD(T)})\).
  \item DLPNO-CCSD(T)/cc-pVDZ, TightPNO - supplies
        \(E(2,\mathrm{TPNO\text{-}CCSD(T)})\).
  \item DLPNO-CCSD(T)/cc-pVTZ, NormalPNO - supplies
        \(E(3,\mathrm{NPNO\text{-}CCSD(T)})\).
  \item RI-MP2/cc-pVTZ - supplies \(E(3,\mathrm{MP2})\) and,
        from its underlying SCF, \(E_{\mathrm{HF}}^{(3)}\).
  \item RI-MP2/cc-pVQZ - supplies \(E_{\mathrm{MP2}}^{(4)}\)
        (hence the \(N=4\) correlation energy) and, from its SCF,
        \(E_{\mathrm{HF}}^{(4)}\).
\end{enumerate}

The scripts associated with these are available in the GitHub repository of this work and carry out this segment by segment computation. Therefore the time we report for the CBS is the cumulative time of each of the individual segments.

\begin{table}[htb!]
\centering
\caption{Computational time cost per fidelity level for the ANI-1ccx dataset.}
\label{tab:ani_time_cost}
\begin{tabular}{|l|c|}
\hline
\textbf{Fidelity Level} & \textbf{Time Cost (s)} \\ 
\hline
HF/ccpVDZ & 9.64615 \\
DFT-$\omega$b97x/6-31G* & 16.8199 \\
DFT-$\omega$b97x/def2-TZVPP & 40.34815 \\
DLPNO-CCSD(T)*/CBS & 157.2737 \\ 
\hline
\end{tabular}
\end{table}

\section{Molecular Descriptors}
Molecular descriptors are a pre-processing step for the ML-QC pipeline and result in featurization of the molecule for the ML model to map to the output property \cite{lilienfeld_2015_descriptors_requirement}. 

In this work, for the VIB5 and QeMFi datasets, unsorted Coulomb Matrices (CM) \cite{RupCM} are used as molecular descriptors or \textit{representations}. 
Consider a molecule with atoms indexed as $i$ with corresponding nuclear charge $Z_i$, The CM is then computed as:
\begin{equation}
CM_{i,j}:=
    \begin{cases}
        \frac{Z_i^{2.4}}{2}~,&i=j\\
        \frac{Z_i\cdot Z_j}{\left\lVert \boldsymbol{R}_i-\boldsymbol{R}_j\right\rVert}~,&i\neq j~,
    \end{cases}
    \label{CM_eq}
\end{equation}
where $\boldsymbol{R}_i$ are the Cartesian coordinates of the atom. Since the CM is a symmetric representation, it is of size $n(n-1)/2$ for a molecule with $n$ atoms.

For the ANI-1ccx dataset, this work uses the SLATM \cite{Huang2020slatm} representation which in our previous experience has worked well for cases where the data consists of different molecules \cite{zasp19a, Vinod2024_multifidelity_monomers_CCSDT}. 

\section{Validation Set and Test Set MAEs}
\begin{figure}[htb!]
    \centering
    \includegraphics[width=0.95\linewidth]{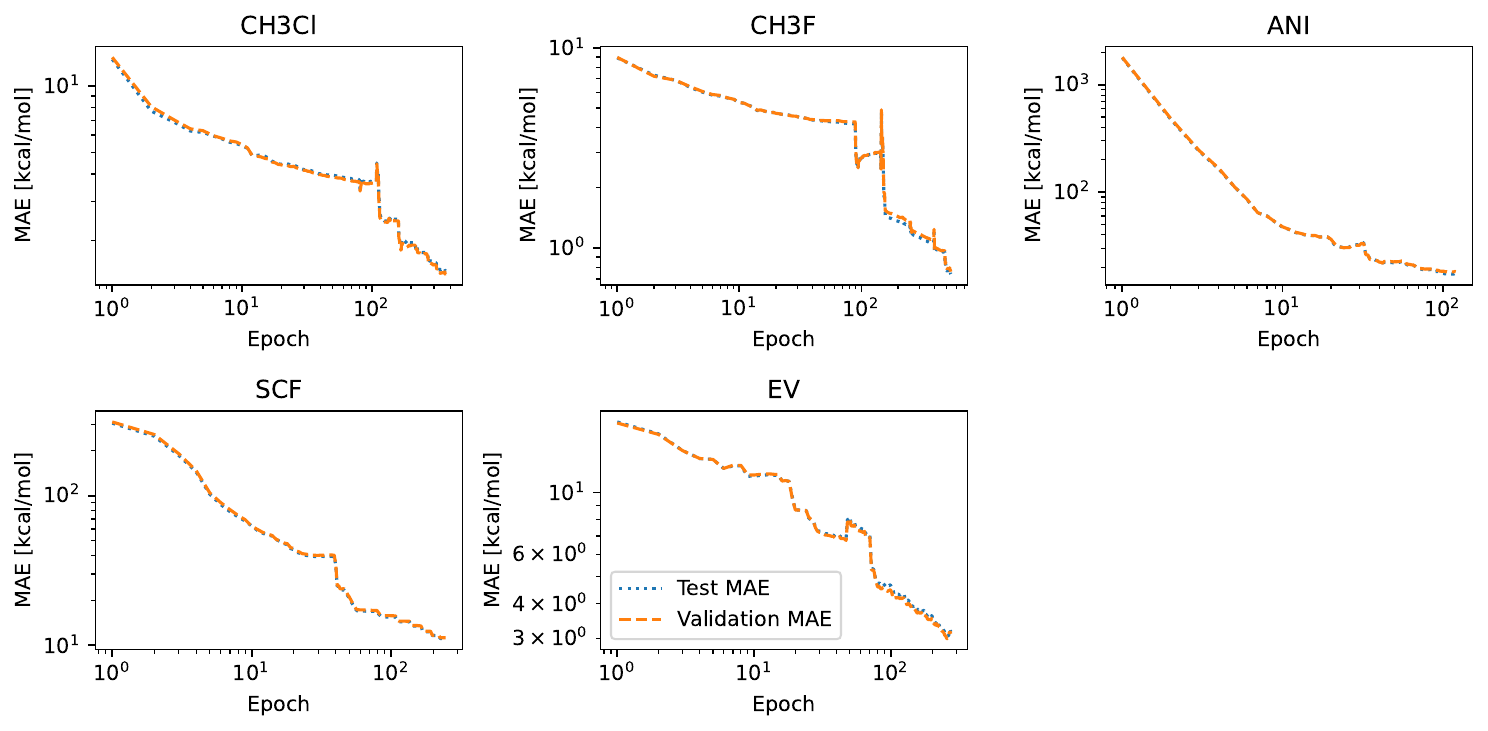}
    \caption{Comparing MAE of the adaptive-MFML model on the validation set and the test set. In the main text, the learning curves are shown for the MAE of the test set.}
    \label{fig_val_test_mae}
\end{figure}
In the main text, the learning curves are shown for the MAE computed on the holdout test set to truly present the empirical error reduction using the method. However, the adaptive-MFML method relies on the validation set, of a much smaller size, to inform addition of training samples at each fidelity. Therefore, it is relevant that for completeness we look at the MAE evolution over the validation set and compare it to that over the evolution for the test set. Fig.~\ref{fig_val_test_mae} shows this comparison for all the datasets/cases studied in the main text. Note that this figure compares the MAEs and not the moving average of the errors which is used in the algorithm. In all cases, the validation MAE closely follows the test MAE indicating that the use of a validation set to inform the adaptive-MFML is indeed a viable option.

\section{ANI Prediction Errors}
\begin{figure}[htb!]
    \centering
    \includegraphics[width=0.85\linewidth]{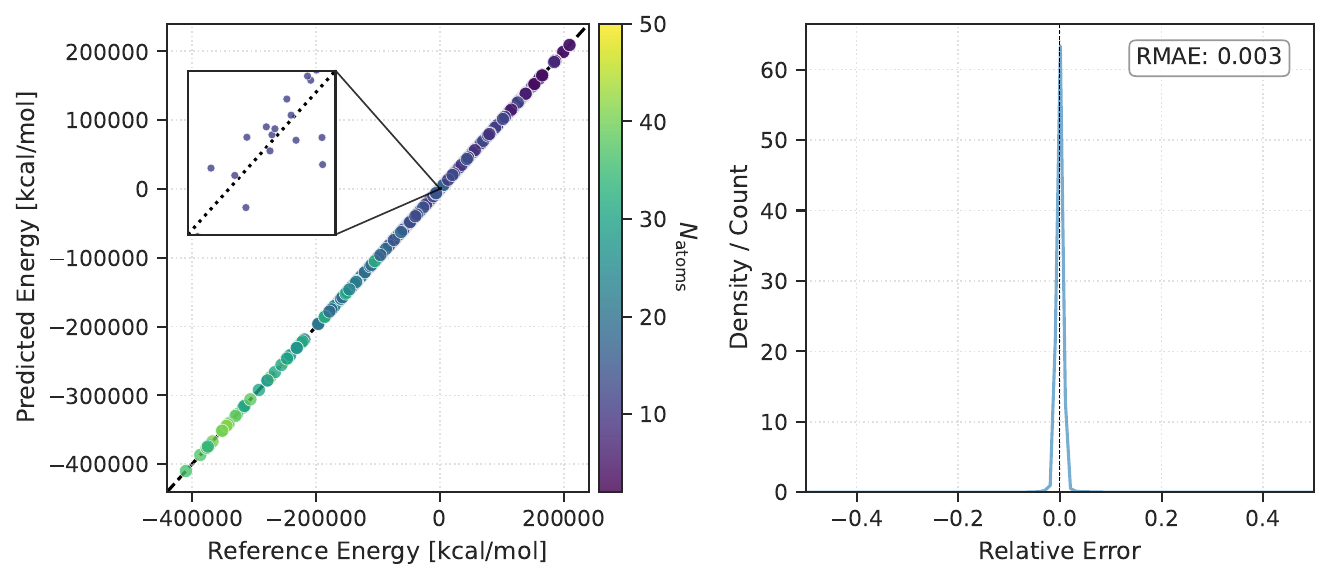}
    \caption{Analyzing the behavior of error of prediction for the ANI-1ccx dataset. The scatter plot of predicted versus reference energies is accompanied by the distribution of the difference in the two for the adaptively sampled MFML model.}
    \label{fig_ANI_scatter}
\end{figure}

In Fig.~\ref{fig_ANI_scatter}, two components of further assessing the resulting error of the adaptively sampled MFML model are provided. Often it is relevant to look not only at the MAE as the metric for model accuracy but to also look at the predictions themselves. 
To this end, having run the adaptive-MFML scheme, we record the number of training samples that were adaptively selected by the scheme. For this specific case the number of training samples chosen are $[768,112,28,12]$, that is 12 DLPNO-CCSD(T)*/CBS training samples (plus the 2 validation samples). This MFML model is trained and thereafter used to predict over the holdout test set of 5,000 geometries. 

On the left-hand-side of Fig.~\ref{fig_ANI_scatter}, the scatter plot of predicted and reference DLPNO-CCSD(T)*/CBS energies are shown color coded by the number of atoms present in the molecule. The predictions lie very much on the identity line (45 degree line, the ideal of prediction=reference) indicating that the model does indeed make correct predictions. Since the scale of the axes are large, we zoom into the range of energies between 30 and 150 kcal/mol. Here we notice that the scatter points are indeed very close to the identity line. 
The right-hand-side of Fig.~\ref{fig_ANI_scatter} shows the distribution of the relative error computed as 
\[
\text{Error}_{rel,i}:=\left(\frac{y_{i}^F - P_{\rm ML}(\boldsymbol{x}_i)}{y_{i}^F}\right)
\]
for the predictions from the adaptive-MFML model and reference DLPNO-CCSD(T)*/CBS energies. The density peaks at 0 trailing off exponentially towards the higher values indicating that the model indeed predicts very close to the reference values. The adaptively sampled MFML model results in a RMAE of 0.003.

To quantify the actual errors instead of the relative errors we also depict the actual errors in Table \ref{tab:ani_accuracy}. Energies are mean-centered. MAE/RMSE are over all test conformers. The per-atom error and RMAE are computed per molecule and averaged over molecules.
In the table, the results are reported for a single seed (42). The 5,000 test conformers correspond to 4,096 distinct molecules, a mean of 1.22 conformers per molecule (identified by canonical SMILES from RDKit connectivity), so only 904 of those molecules (22\%) contain more than one conformer and a within-molecule energy difference can be formed at all.
\begin{table}[htb!]
  \centering
  \caption{Accuracy of the adaptive-MFML model and of the single-fidelity KRR model on the holdout test set of ANI-1ccx. Both models are evaluated at comparable total acquisition costs ($\sim$3~h).}
  \label{tab:ani_accuracy}
  \begin{tabular}{|l|c|c|c|c|}
    \hline
    \textbf{Model} 
      & \makecell{\textbf{MAE} \\ \textbf{[kcal/mol]}} 
      & \makecell{\textbf{RMSE} \\ \textbf{[kcal/mol]}} 
      & \makecell{\textbf{MAE/atom} \\ \textbf{[kcal/mol]}} 
      & \textbf{RMAE} \\
    \hline
    \hline
    Single-fidelity KRR ($2^6$) & 288.038 & 405.933 & 22.375 & 0.0477 \\
    Adaptive-MFML [768,112,28,12] & 26.042 & 36.872 & 1.867 & 0.0033 \\
    \hline
  \end{tabular}
\end{table}

\section{Ceiling-Adaptive MFML Data Sampling}

\begin{figure}[htb!]
    \centering
    \includegraphics[width=0.95\linewidth]{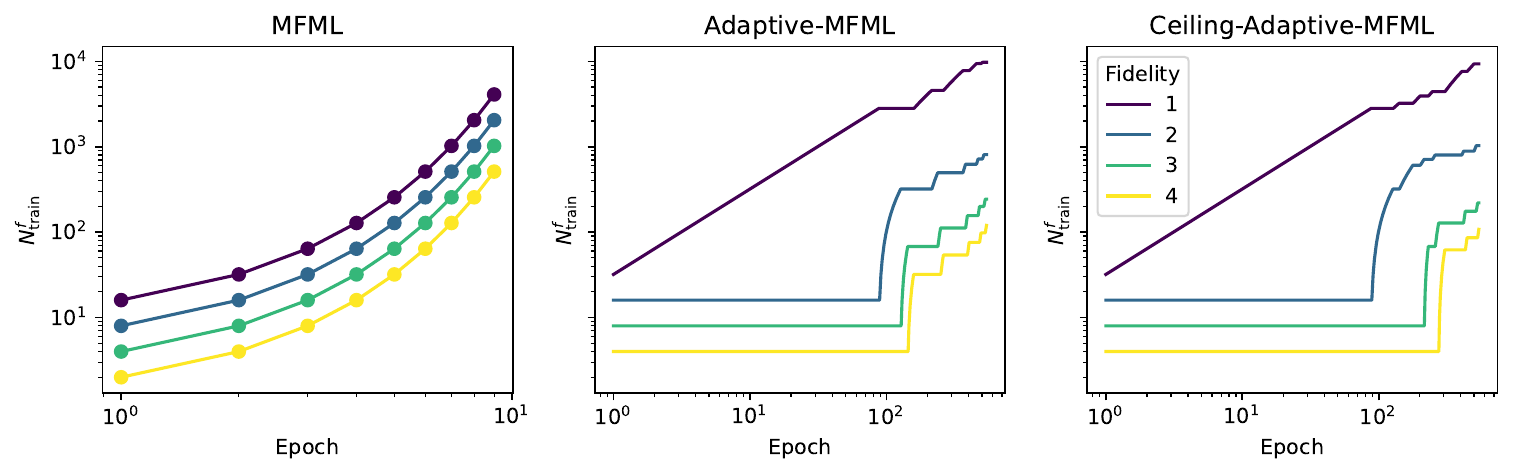}
    \caption{Evolution of training set sizes at different fidelities for basic MFML, adaptive MFML, and ceiling adaptive MFML schemes.}
    \label{fig_train_size_evolution}
\end{figure}
\begin{algorithm}[!htbp]
\caption{Ceiling-Adaptive MFML algorithm. Let $E_k$ be the validation MAE after the $k$-th batch addition and $\bar{E}_k:=\frac{1}{w}\sum_{j=k-w+1}^{k}E_j$ its moving average over a window of $w$ iterations. The local improvement is $\Delta_k:=|\bar{E}_k-\bar{E}_{k-1}|$ and the global improvement is $\Delta_{\rm global}:=E^{(p-1)}-E^{(p)}$ over one pass through all fidelities. Red text marks the differences from Algorithm~1.}
\label{alg:ceiling_adaptive_mfml}
\begin{algorithmic}[1]
\Require Initial Training Set Sizes $\boldsymbol{N}:=[N_1,N_2,\ldots,N_F]$, 
batch sizes $\boldsymbol{m}:=[m_1,m_2,\ldots,m_F]$, 
local tolerances $\boldsymbol{\tau_{\rm local}}$, 
global tolerances $\boldsymbol{\tau_{\rm global}}$, validation dataset $\mathcal{V}^F$
\Ensure QC calculator $g^f_{\rm QC}~,\forall f\in\{1,\ldots,F\}$

\State $\mathcal{T}^f\gets\left\{(x_i,g_{QC}^f(x_i))\right\}_{i=1}^{N_f},~\forall f\in\{1,\ldots,F\}$
\State $\mathcal{T}_{\rm MFML}\gets\left[T^1,\ldots,T^F\right]$
\State $E \gets \text{MAE}(\mathcal{T}_{\rm MFML},\mathcal{V}^F)$
\State $E_{\rm prev} \gets E$
\State $\Delta_{\rm global}\gets E$ \Comment{Force at least one global pass}
\State $f_{\rm end} \gets 2$ \Comment{\textcolor{red}{Initialize ceiling to the second lowest fidelity}}
\While{$\Delta_{\rm global} > \boldsymbol{\tau_{\rm global}}[F]$} \Comment{Ceiling-Adaptive Loop}
    \For{$f = 1$ \textbf{to} $f_{\rm end}$} \Comment{Saturate fidelities up to the current ceiling}
    \State $E \gets \text{MAE}(\mathcal{T}_{\rm MFML},\mathcal{V}^F)$
    \State $\Delta \gets E$ \Comment{Re-initialise local statistic each fidelity loop}
    \While{$\Delta>\boldsymbol{\tau_{\rm local}}[f]$}
    \If{$f>1$ and $2\times(\boldsymbol{N_{\rm train}}[f] + \boldsymbol{m_{\rm batch}}[f]) \ge \boldsymbol{N_{\rm train}}[f-1]$}
            \State \textbf{break} \Comment{Maintain hierarchical size constraint}
        \EndIf
    \If{$f>1$}
            \State $\mathcal{X}^f\gets \{x:(x,y)\in\mathcal{T}^{f-1}\}\setminus\mathcal{E}^f$     \Comment{Ensure nested MF data, each geometry queried once}
        \EndIf
        \State $\mathcal{T}^f_{\rm batch}\gets \left\{(\boldsymbol{x}_i,g^f_{\rm QC}(\boldsymbol{x}_i))\right\}_{i=1}^{\boldsymbol{m_{\rm batch}}[f]} \text{, s.t.}~x_i\in\mathcal{X}^f$ \Comment{\textbf{On-the-fly computations}}
        \State remove $\{\boldsymbol{x}_i\}$ from $\mathcal{X}_{\rm pool}$ \Comment{Each geometry is queried once}
        \State $\mathcal{T}^f\gets\mathcal{T}^f\cup\mathcal{T}^f_{\rm batch}$ \Comment{Add samples at current fidelity}
        \State $\mathcal{E}^f\gets\mathcal{E}^f\cup\{\boldsymbol{x}_i\}$ \Comment{Update evaluated set}
        \State $\mathcal{T}_{\rm MFML} \gets \left[\mathcal{T}^1,\ldots \mathcal{T}^f,\ldots,\mathcal{T}^F\right]$ \Comment{Update multifidelity training data}
        \State $N_{\rm train}[f] \gets \boldsymbol{N_{\rm train}}[f] + \boldsymbol{m_{\rm batch}}[f]$ \Comment{Update training set size}
        \State $E \gets \text{MAE}(\mathcal{T}_{\rm MFML},\mathcal{V}^F)$
        \State $\Delta \gets |\bar{E}_k-\bar{E}_{k-1}|$ \Comment{Local improvement}
    \EndWhile
    \EndFor
    \State $\Delta_{\rm global}\gets E_{\rm prev}-E$ \Comment{Global improvement} \State $E_{\rm prev}\gets E$
    \If{$f_{\rm end}<F$ and $\Delta_{\rm global} < \boldsymbol{\tau_{\rm global}}[f_{\rm end}]$}
        \State $f_{\rm end} \gets f_{\rm end} + 1$ \Comment{\textcolor{red}{Raise the fidelity ceiling if improvement stagnates}}
    \EndIf
\EndWhile
\State \Return $\boldsymbol{N_{\rm train}}, \mathcal{T}^f, \forall f\in\{1,2,\ldots,F\}$
\end{algorithmic}
\end{algorithm}

In addition to the adaptive sampling scheme reported in the main text, an additional sampling scheme was assessed for this work. It uses the same definitions of the local and global improvements and the same single-use pool as Algorithm~1 of the main text (the local statistic is re-initialised at the start of every fidelity loop). The Ceiling-Adaptive Sampling algorithm is presented in Algorithm \ref{alg:ceiling_adaptive_mfml} with key differences to the Adaptive-MFML algorithm from Algorithm~1 being highlighted in red. The algorithm for iterations over fidelities is slightly modified. As before, we begin with a small basic MFML model, say, $[32,16,8,4]$. Instead of adding training samples across all fidelities, the scheme now add samples only at the first two fidelities, that is sets the active fidelity ceiling at $f=2$. Once  a global tolerance level for error improvement is crossed the active ceiling of fidelities is moved from $f=2$ to $f=3$. Now the scheme recursively adds samples at $f\leq3$ fidelities until the global tolerance criterion is crossed and the active fidelity ceiling is updated to $f=4$. This process is repeated until the highest fidelity is reached. At this stage, notice that the ceiling-adaptive scheme has returned to the Adaptive MFML scheme mentioned in the main text. The difference in evolution of training data set sizes at different fidelities for the basic MFML scheme, Adaptive-MFML, and Ceiling-Adaptive MFML are shown in Fig.~\ref{fig_train_size_evolution}. Notice that for the Ceiling-Adaptive MFML, training samples are added the highest fidelity only to the very end of the scheme since the algorithm spends quite some time attempting improvement at the lower fidelities first. 

\begin{figure}[htb!]
    \centering
    \includegraphics[width=1\linewidth]{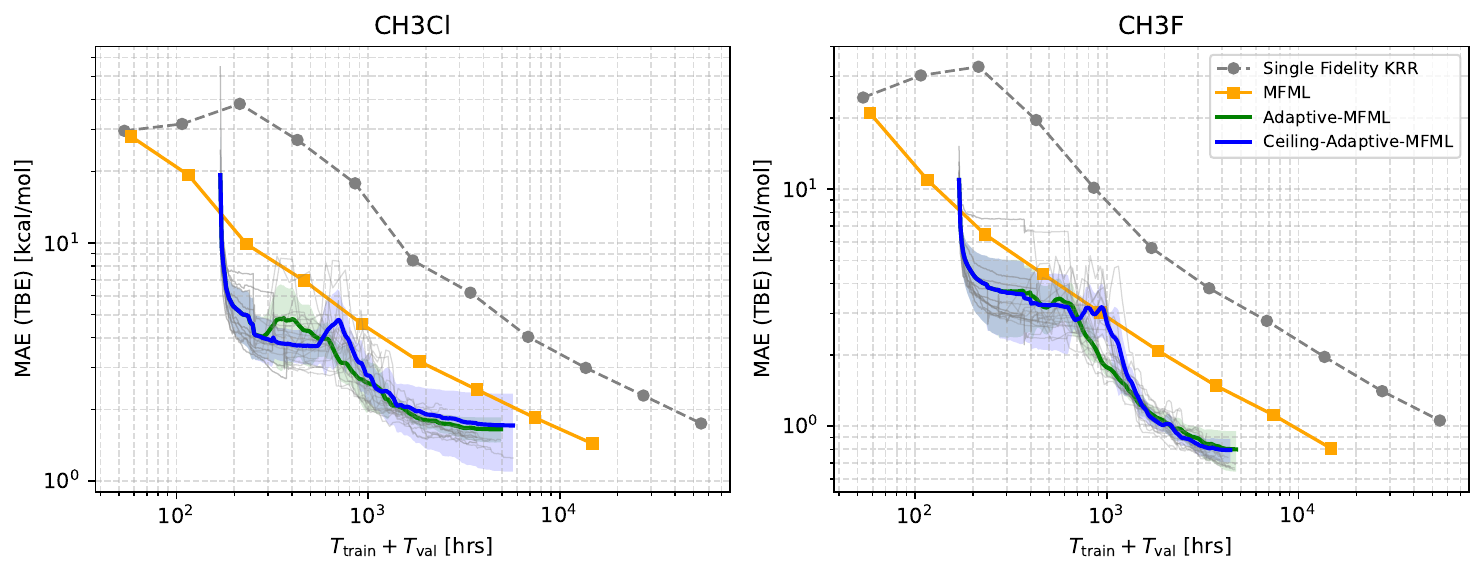}
    \caption{Time-cost of training data versus MAE for prediction of \textit{ab initio} potential energy surfaces at the TBE for two molecules of VIB5 \cite{zhang_vib5_2022}.}
    \label{fig_VIB5_ceiling_LC}
\end{figure}
 
\begin{figure}[htb!]
    \centering
    \includegraphics[width=1\linewidth]{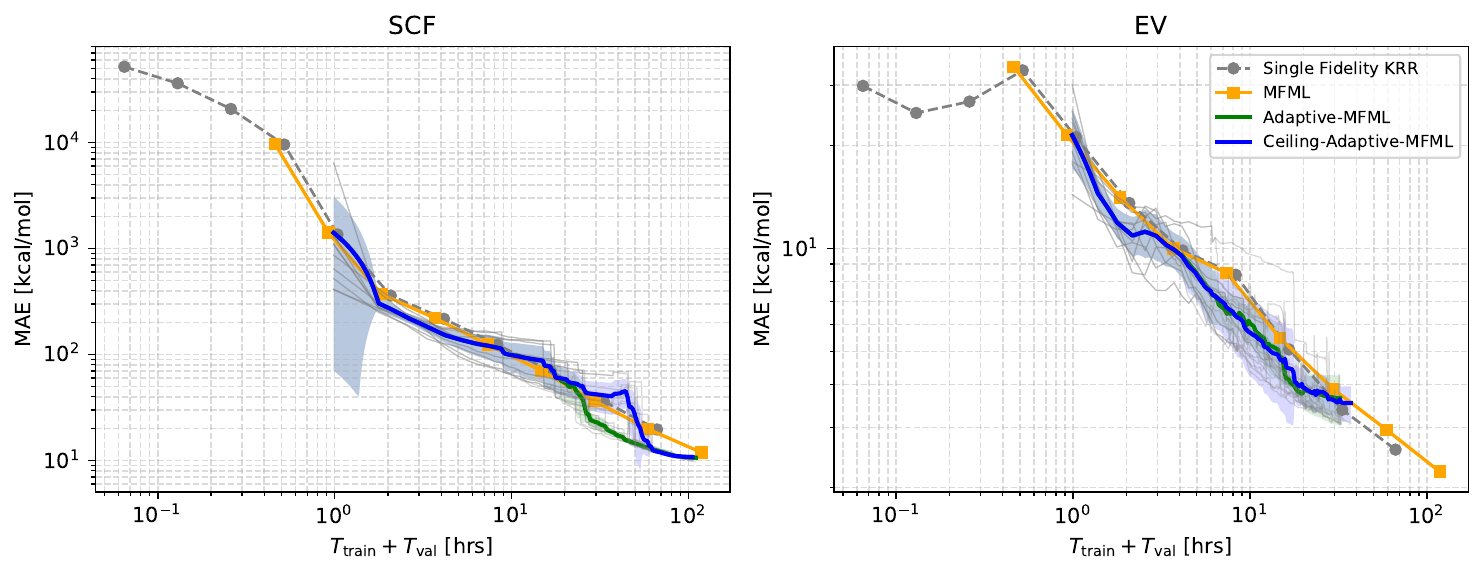}
    \caption{Assessing the model error as a function of time-cost incurred in generating training data for single fidelity KRR, MFML, and ceiling adaptive-MFML on the QeMFi dataset \cite{vinod2024QeMFi_paper, vinod_2024_QeMFi_zenodo_datatset}.}
    \label{fig_QeMFi_ceiling_LC}
\end{figure}

\begin{figure}[htb!]
    \centering
    \includegraphics[width=0.65\linewidth]{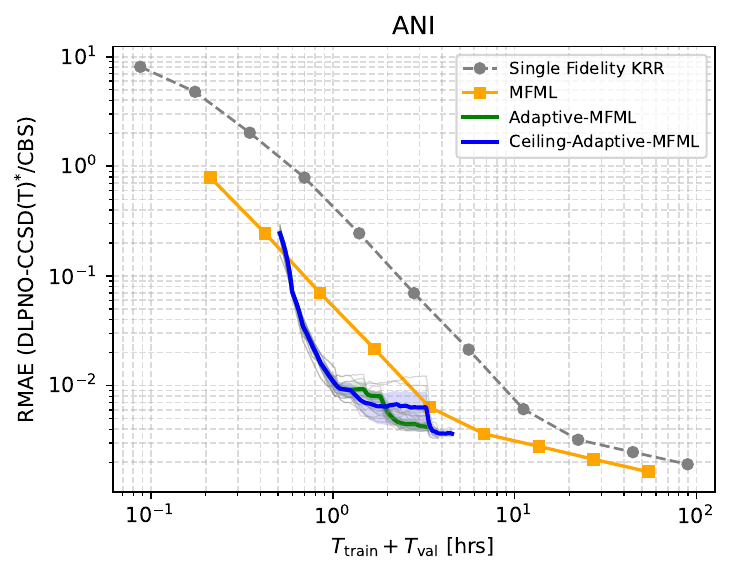}
    \caption{Time-cost of training data versus MAE for the ANI1-ccx dataset collection of molecules \cite{2020_Smith_ANI1ccx_dataset}.}
    \label{fig_ANI_ceiling_LC}
\end{figure}

Time versus MAE curves for the Ceiling-Adaptive MFML are shown in Fig.~\ref{fig_VIB5_ceiling_LC} for molecules of VIB5, Fig.~\ref{fig_QeMFi_ceiling_LC} for the SCF and EV properties of QeMFi, and Fig.~\ref{fig_ANI_ceiling_LC} for the ANI-1ccx dataset.

Consider the case for the ANI-1ccx dataset as seen in Fig.~\ref{fig_ANI_ceiling_LC} where the Ceiling-adaptive MFML is compared to the other methods studied in the main text. It is seen that the MAE curve stagnates for x-axis values between 1.5 and 5 hrs of time-cost. Following this, the error once again starts to drop and reaches near the Adaptive-MFML MAE values. This behavior can be understood as follows: the Ceiling-Adaptive-MFML only adds training samples up to the active ceiling fidelity. The highest fidelity is only `visited' by the algorithm after several epochs of adding training data at the lower fidelities. This results in the MFML model not having enough information about the higher fidelities until much later in the scheme. This corresponds to the sudden drop in the MAE that we observe in Fig.~\ref{fig_ANI_ceiling_LC} around $\sim~4$~hrs of time-cost. Eventually, the Ceiling-Adaptive-MFML settles down to the Adaptive-MFML scheme altogether since the active ceiling fidelity is the top fidelity.

\section{Local and Global Tolerance Grid}\label{sec:tolerance_grid}
\begin{table}[htb!]
    \centering
    \begin{tabular}{|c|c|c|c|}
         \hline
         \textbf{Dataset} & \textbf{Local Tolerance} & \textbf{Global Tolerance} & \textbf{Window Length} \\
         \hline
         \hline
         VIB5 ($\rm CH_3Cl$)& $[10^{-2},10^{-2},10^{-2},10^{-2}]$& $10^{-1}$& 10 \\
         VIB5 ($\rm CH_3F$)& $[10^{-3},10^{-3},10^{-3},10^{-3}]$& $10^{-1}$& 10 \\
         QeMFi (SCF)& $[1,1,2,2]$& $0.5$& 5 \\
         QeMFi (EV)& $[10^{-1},10^{-1},10^{-1},10^{-1}]$& $10^{-1}$& 5 \\
         ANI & $[1,1,1,1]$& 1& 5 \\
         \hline
    \end{tabular}
    \caption{Parameters for the adaptive-MFML scheme for each of the dataset studied in the main text.}
    \label{tab_adaptiveMFML_params}
\end{table}
The Adaptive MFML scheme introduces further parameters into the overarching model. This involves the local and global tolerance values, and the batch sizes of addition of training samples at each fidelity. Additionally, the window length of the MAE where the improvement of the model is computed is also a hyper-parameter.  
For the experiments in the main text, these values were manually set. No optimization was carried out for these. 
The value of hyper-parameters are reported in Table \ref{tab_adaptiveMFML_params} for reproducibility.

At the same time, we wish to show that the variation of these introduced hyperparameters is not significant on the final MAE of the adaptive-MFML model. To assess the sensitivity of the adaptive-MFML stopping rule to its two tolerance hyperparameters, we evaluated a $5\times5$ grid with local and global tolerances $\tau,\tau_{\rm global}\in\{0.01,0.1,1,2,4\}$ at $N_{\rm val}=2$.

\begin{figure}[htb!]
    \centering
    \includegraphics[width=\linewidth]{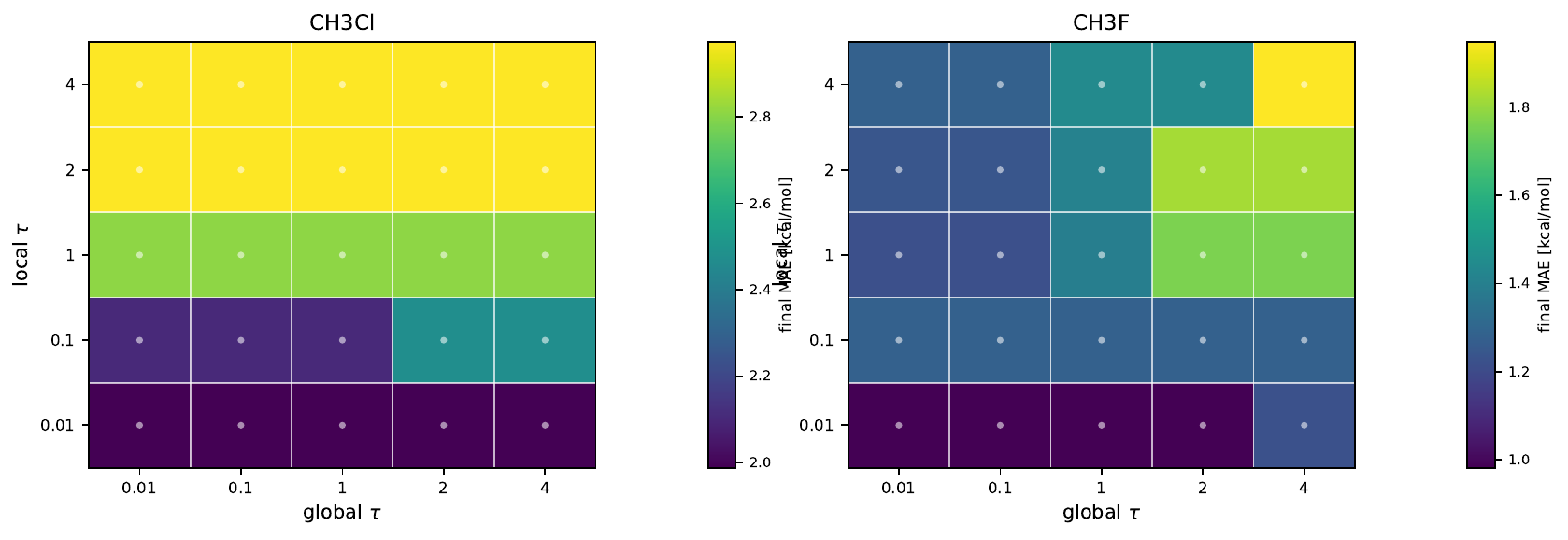}
    \caption{Final test MAE over the $5\times5$ local/global tolerance grid for CH$_3$Cl and CH$_3$F of the VIB5 dataset.}
    \label{fig:tol_grid_vib5}
\end{figure}

\begin{figure}[htb!]
    \centering
    \includegraphics[width=\linewidth]{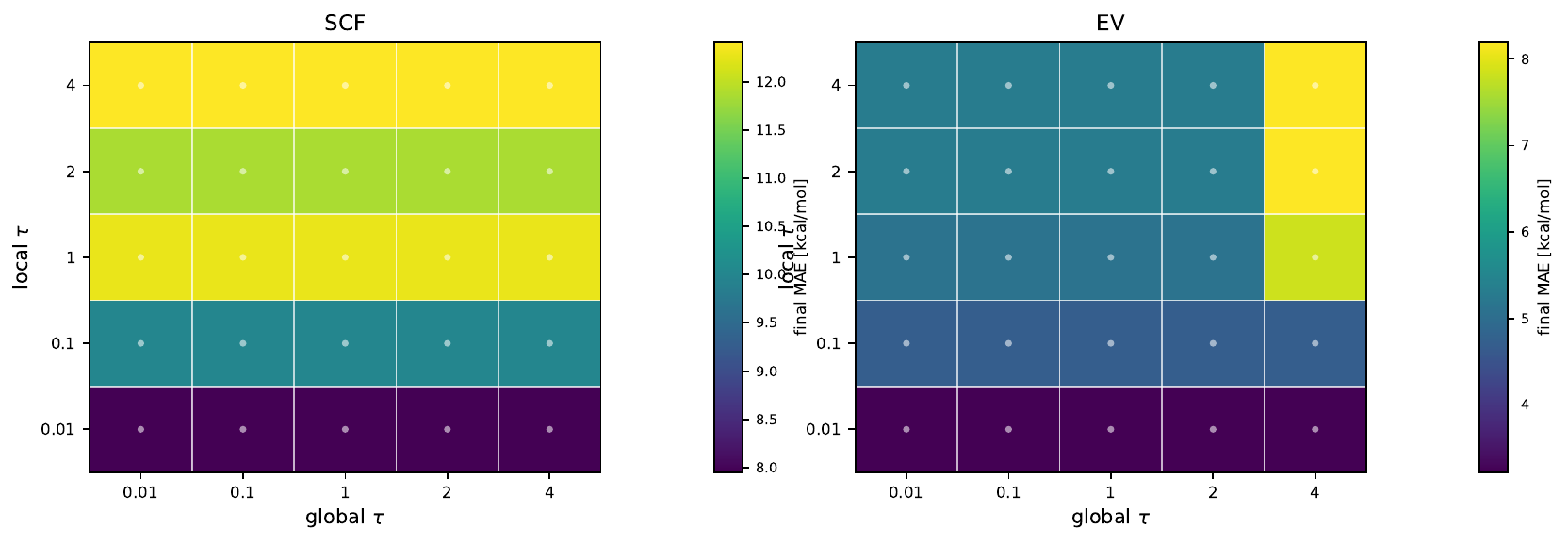}
    \caption{Final test MAE over the local/global tolerance grid for the SCF and EV properties of the QeMFi dataset.}
    \label{fig:tol_grid_qemfi}
\end{figure}

\begin{figure}[htb!]
    \centering
    \includegraphics[width=0.65\linewidth]{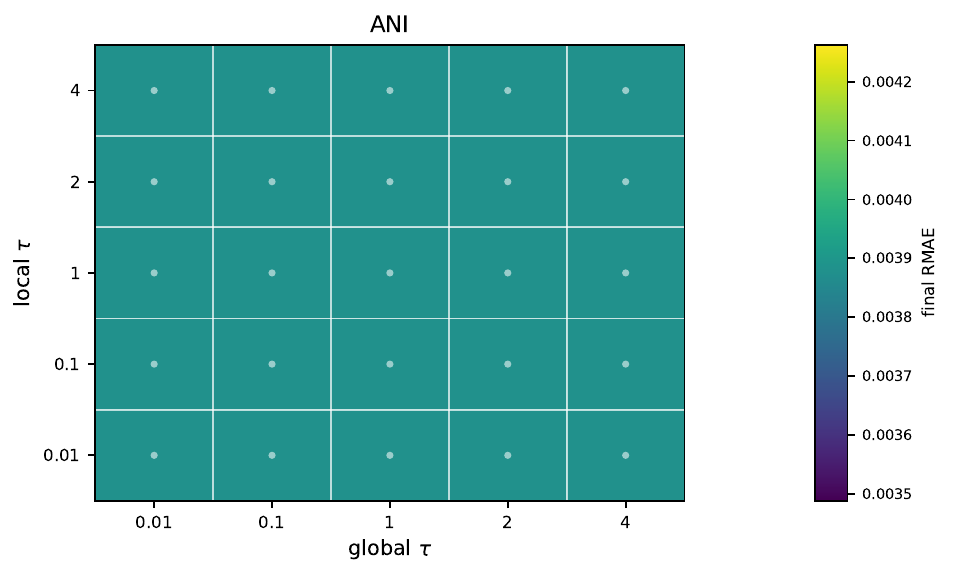}
    \caption{Final test RMAE over the local/global tolerance grid for
    ANI-1ccx. The vertical axis, reports RMAE, and is scaled by $10^{3}$.}
    \label{fig:tol_grid_ani}
\end{figure}

Figures~\ref{fig:tol_grid_vib5}--\ref{fig:tol_grid_ani} show the final test error over the grid. The local tolerance appears have a more dominant control. For a fixed global tolerance, the final error rises from 2.0 to 2.85~kcal/mol for CH$_3$Cl, from 1.0 to 1.85~kcal/mol for CH$_3$F, from 8.0 to 12.0~kcal/mol for QeMFi SCF and from 4.0 to 8.0~kcal/mol for QeMFi EV as $\tau_{\rm
local}$ is loosened. The global tolerance has little effect within a row, except where the local tolerance is already loose (QeMFi EV at $\tau_{\rm local}=4$, where the final error rises from 6 to 8~kcal/mol between $\tau_{\rm global}=0.01$ and 4). The ANI-1ccx result is
essentially uniform over the whole grid ($3.5$--$4.2\times10^{-3}$ in RMAE). Every grid point is a single-seed run.
The tolerances used for the main-text experiments (Table~\ref{tab_adaptiveMFML_params}) sit in the loose part of the grid for QeMFi SCF ($\tau_{\rm local}=1$--2), which is why that run terminates at $\sim$13~kcal/mol. For VIB5 they are in the tightest row ($\tau_{\rm local} =0.01$), consistent with the 2~kcal/mol target used there.

\section{Effect of the Validation Set Size}\label{sec:validation_cost}

The adaptive-MFML stopping rule evaluates the validation error after every batch addition, so the high-fidelity labels of the validation set are generated once and add a fixed offset to the training cost. Table~\ref{tab:validation_cost} lists the cost of a single highest-fidelity label and the resulting validation cost for the validation sizes considered in this work. The main text uses the smallest practical set, $N_{\rm val}=2$.

\begin{table}[htb!]
  \centering
  \caption{Cost of the validation labels. $c_F$ is the cost of one highest-fidelity reference calculation, $N_{\rm val}^{(5\%)}$ is the size of the full 5\,\% validation split we make of the datasets, and the last two columns give the total validation cost for the full split and for $N_{\rm val}=2$.}
  \label{tab:validation_cost}
  \begin{tabular}{|l|c|c|c|c|}
    \hline
    \textbf{Dataset} & \textbf{$c_F$ [h]} & \textbf{$N_{\rm val}^{(5\%)}$} 
    & \textbf{cost, 5\,\% [h]} & \textbf{cost, $N_{\rm val}=2$ [h]} \\
    \hline
    \hline
    VIB5 (CH$_3$Cl)  & 26.71 & 2,241 & 59,850 & 53.4 \\
    VIB5 (CH$_3$F)   & 26.71 & 4,133 & 110,379 & 53.4 \\
    QeMFi            & 0.101 & 6,750 & 681.8 & 0.2 \\
    ANI-1ccx         & 0.0437 & 2,500 & 109.2 & 0.1 \\
    \hline
  \end{tabular}
\end{table}

\begin{figure}[htb!]
  \centering
  \includegraphics[width=\linewidth]{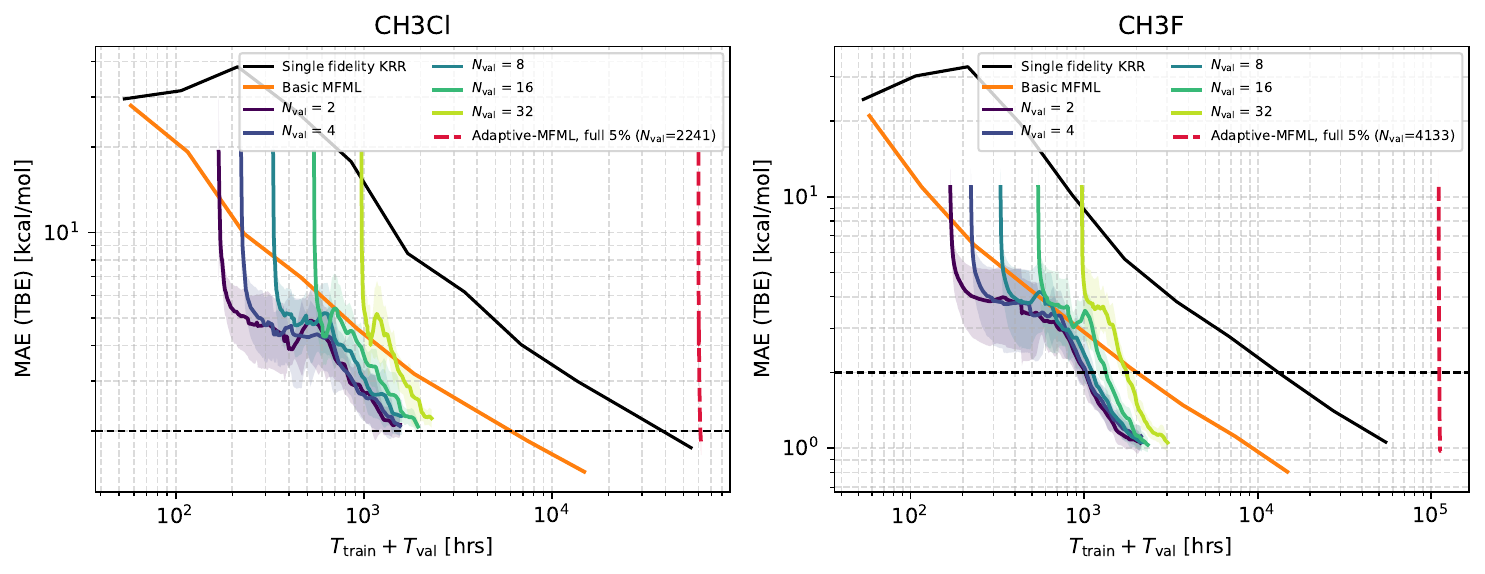}
  \caption{CH$_3$Cl and CH$_3$F: test MAE versus total cost (training plus
  validation labels) for single-fidelity KRR, basic MFML, and adaptive-MFML
  with $N_{\rm val}=2,4,8,16,32$.}
  \label{fig:validation_sweep_vib5}
\end{figure}

\begin{figure}[htb!]
  \centering
  \includegraphics[width=\linewidth]{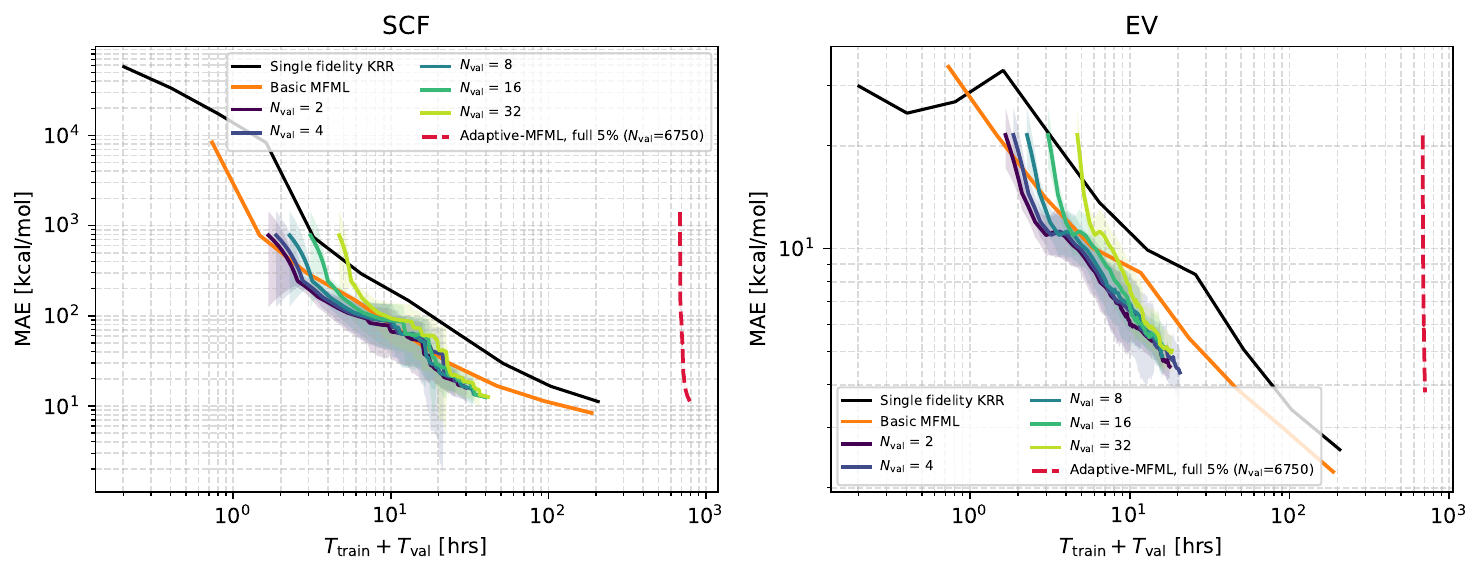}
  \caption{QeMFi SCF and EV: test MAE versus total cost for single-fidelity KRR, basic MFML, and adaptive-MFML with $N_{\rm val}=2,4,8,16,32$.}
  \label{fig:validation_sweep_qemfi}
\end{figure}

\begin{figure}[htb!]
  \centering
  \includegraphics[width=0.65\linewidth]{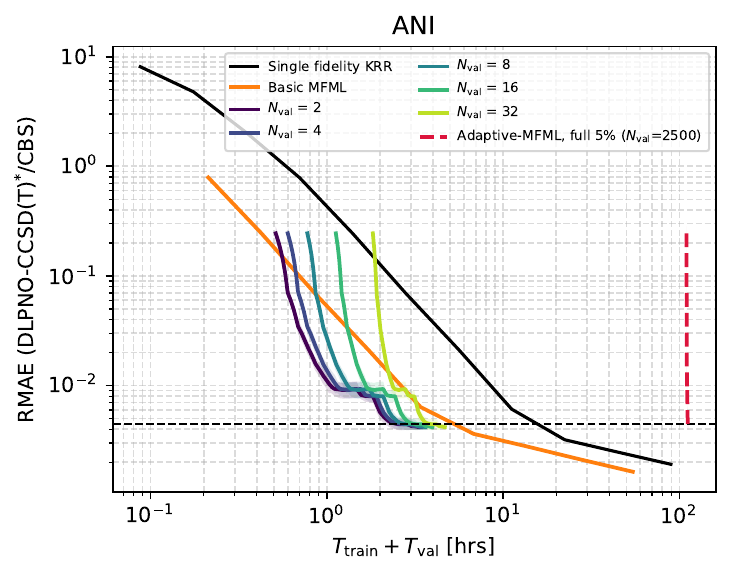}
  \caption{ANI-1ccx: test RMAE versus total cost for single-fidelity KRR, basic MFML, and adaptive-MFML with $N_{\rm val}=2,4,8,16,32$.}
  \label{fig:validation_sweep_ani}
\end{figure}

Figures~\ref{fig:validation_sweep_vib5}--\ref{fig:validation_sweep_ani}
compare the learning curves for $N_{\rm val}=2,4,8,16,32$ (ten seeds each), plotted against the total cost $T=T_{\rm train}+N_{\rm val}c_F$. The single-fidelity KRR and basic-MFML curves do not use a validation set and are shown for reference at training cost only. The tolerances and batch sizes are those of Table~\ref{tab_adaptiveMFML_params}. For VIB5, where a single TBE label costs 26.71 h, setting
$N_{\rm val}=2$ removes a factor of 16 in terms of cost with respect to the 32 validation set sizes, and by roughly three orders of magnitude in comparison with the full $5\%$ split (1,121-fold for CH$_3$Cl and 2,067-fold for CH$_3$F). The final test error is essentially unchanged over this range. The MAE varies by at most 0.07 kcal/mol for CH$_3$Cl and 0.13 kcal/mol for CH$_3$F, while the total cost to reach 2 kcal/mol drops from 2,464 h to 1,378 h for CH$_3$Cl and from 1,704 h to 961 h for CH$_3$F. 

For QeMFi, the labels are cheap (0.10 h each), the validation cost stays below 4 h at $N_{\rm val}=32$, and the final error shows no consistent trend with $N_{\rm val}$. SCF varies by up to 2.2 kcal/mol across the different choices of $N_{\rm val}$, and for EV by 0.6 kcal/mol. 

For ANI-1ccx, the adaptive trajectory is identical for every $N_{\rm val}$ to within $10^{-10}$ with only the cost offset changing. The run terminates at a total cost of 2.46~h at $N_{\rm val}=2$ versus 3.77~h at $N_{\rm val}=32$, a difference of $30\times0.0437=1.31$~h.

\newpage

\end{document}